\documentclass[11pt]{article}

\usepackage[preprint]{acl}

\usepackage{times}
\usepackage{latexsym}

\usepackage[T1]{fontenc}

\usepackage[utf8]{inputenc}

\usepackage{microtype}

\usepackage{inconsolata}

\usepackage{graphicx}

\usepackage{times}
\usepackage{latexsym}
\usepackage{graphicx}
\usepackage{xcolor}
\usepackage[skip=2pt]{caption}
\usepackage[hang,flushmargin]{footmisc}
\usepackage{amssymb}
\usepackage{booktabs}
\usepackage{multirow}
\usepackage{xcolor}
\usepackage[most]{tcolorbox}
\tcbset{enhanced, breakable, colback=white, colframe=black, boxrule=0.4pt}
\usepackage{lipsum}
\usepackage{amsmath}
\usepackage{tikz}
\usepackage{subcaption}
\usepackage{soul}
\usepackage[normalem]{ulem}
\usepackage[table]{xcolor}
\usepackage{colortbl}
\usepackage{hhline}
\usepackage{makecell}
\usepackage{multirow}
\usepackage{tabularx}
\usepackage[linesnumbered, ruled, vlined]{algorithm2e}
\usepackage{placeins}

\title{Hierarchical Multi-Persona Induction from User Behavioral Logs: Learning Evidence-Grounded and Truthful Personas}

\author{
 \textbf{Nayoung Choi\textsuperscript{1}},
 \textbf{Haeyu Jeong\textsuperscript{2}},
 \textbf{Changbong Kim\textsuperscript{2}},
 \textbf{Hongjun Lim\textsuperscript{2}},
 \textbf{Jinho D. Choi\textsuperscript{1}}
\\
 \textsuperscript{1}Department of Computer Science, Emory University, Atlanta, GA, USA\\
 \textsuperscript{2}Naver Corporation, South Korea
\\
 \small{
 \texttt{\{nayoung.choi, jinho.choi\}@emory.edu}, \texttt{\{haeyu.jeong, changbong.kim, hongjun.lim\}@navercorp.com}
 }
}

\begin{document}
\maketitle

\begin{abstract}
Behavioral logs provide rich signals for user modeling, but are noisy and interleaved across diverse intents. Recent work uses LLMs to generate interpretable natural-language personas from user logs, yet evaluation often emphasizes downstream utility, providing limited assurance of persona quality itself. We propose a hierarchical framework that aggregates user actions into intent memories and induces multiple evidence-grounded personas by clustering and labeling these memories. We formulate persona induction as an optimization problem over persona quality—captured by cluster cohesion, persona–evidence alignment, and persona truthfulness—and train the persona model using a groupwise extension of Direct Preference Optimization (DPO). Experiments on a large-scale service log and two public datasets show that our method induces more coherent, evidence-grounded, and trustworthy personas, while also improving future interaction prediction.



\end{abstract}

\section{Introduction}
Large-scale behavioral logs are continuously accumulated by modern search and recommendation systems. These logs provide a detailed record of what users query and click, and thus offer a valuable substrate for user modeling. Better user representations help these systems personalize user experiences over time \cite{liu2023pre, peng2025survey, xu2025personalized}. However, behavioral logs are also inherently noisy, fragmented, and interleaved across contexts: a single user may alternate between short-lived goals and long-term preferences, with multiple intents mixed within the sequence. This heterogeneity calls for modeling behavior at an appropriate level of abstraction, so that stable user characteristics can be captured without conflating distinct contexts.

\begin{figure*}[htp!]
    \centering
    \includegraphics[width=1.0\linewidth]{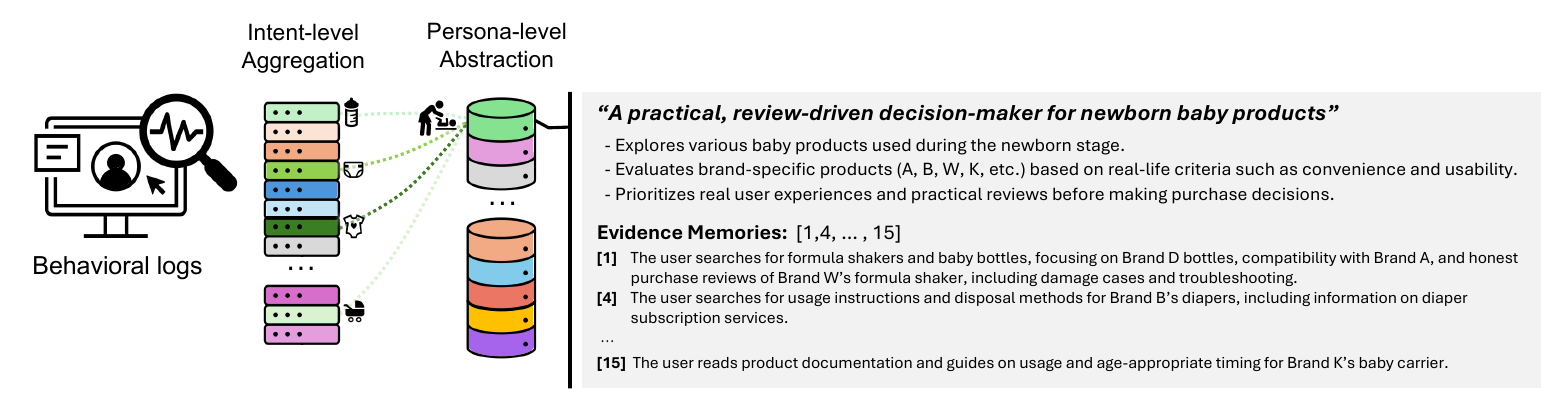}
    \caption{Overview of our hierarchical multi-persona induction framework. Behavioral logs of a user are aggregated into \textit{intent memories} and abstracted into evidence-grounded \textit{user personas}. The persona model is trained with rewards defined over three primary quality criteria: cluster cohesion, persona–evidence alignment, and persona truthfulness. The right panel shows an example persona ($p_i$) induced from \texttt{Srv.} logs using our pipeline.}
    \label{fig:overview}
    \vspace{-0.2cm}
\end{figure*}

\noindent Recently, large language models (LLMs) have been used to produce natural-language personas from user logs, providing an interpretable view of users’ behavioral patterns and preferences. Such personas offer a human-readable form of user representation, facilitating their use in various downstream applications. Despite these advantages, a key question remains: \textit{How can we define and evaluate persona quality?} Most prior work evaluates or optimizes personas primarily through downstream utility such as user behavior prediction \cite{chen2025deeper, shi2025personax, wang2025lettingo, gao2025langptune}. While this focus is practically important, downstream performance alone provides limited assurance that the personas are coherent, trustworthy abstractions of the underlying user logs. As a result, persona quality itself remains underexplored.

To address this gap, we propose a hierarchical framework for multi-persona induction from noisy behavioral logs that prioritizes persona quality. Starting from raw user actions, we aggregate contiguous actions with the same intent into \textit{intent memories}. We then cluster and label these memories to construct \textit{user personas}, where each persona represents a distinct user characteristic supported by an evidence set of memories (Figure~\ref{fig:overview}). We formulate persona induction as an optimization problem over persona quality—captured by cluster cohesion, persona–evidence alignment, and persona truthfulness—and design reward functions based on these criteria. Using a groupwise extension of Direct Preference Optimization (DPO), we train the persona model to produce coherent, evidence-grounded, and truthful personas. We further show that optimizing persona quality also improves downstream utility, as measured by future interaction prediction. Our contributions can be summarized as follows:
\vspace{-0.15cm}
\begin{itemize}
\setlength\itemsep{-0.1cm}
    \item \textbf{Evidence-traceable persona induction.} We propose a hierarchical framework that constructs multiple natural-language personas per user, each paired with an explicit evidence set of supporting intent memories.
    \item \textbf{Formalizing persona quality.} We define persona quality using three criteria—cohesion, alignment, and truthfulness—and turn them into learning signals for persona induction.
    \item \textbf{Quality--utility alignment.} We empirically show that optimizing for persona quality under these criteria also improves downstream future prediction across datasets and models.
\end{itemize}


\section{Related Work}

\paragraph{User Modeling from Behavioral Logs} Learning user representations from historical interactions is a central theme in user modeling and personalization, where systems encode behavioral traces---such as queries, clicks---into latent representations to support downstream tasks. A large body of work studies how to represent users from behavioral logs, including sequential and session-based modeling that captures evolving interests over time \cite{10.1145/3477495.3531910, yang-liang-2025-multi}, as well as approaches that aggregate heterogeneous signals across actions and domains \cite{qi-etal-2021-hierec, li-etal-2022-miner}. These methods have been effective for predicting future behavior, but their representations are often implicit and not directly interpretable. Motivated by the observation that users exhibit multiple behavioral facets, a line of works have explored multi-interest or multi-facet representations that decompose a user's history into several components rather than a single representation \cite{10.1145/3394486.3403344, qi-etal-2021-uni-fedrec, zheng-etal-2024-mitigating}.

\paragraph{Natural-Language User Personas} Recent work has increasingly represented users with natural-language personas, motivated by their interpretability and flexibility compared to latent embeddings. Prior studies argue that natural-language user profiles can make personalization more transparent and scrutable, and can be edited to control system behavior \cite{10.1145/3331184.3331211, 10.1145/3477495.3531873, ramos-etal-2024-transparent}. Generating compact textual profiles from personal context has been shown to improve personalization across tasks as an explicit user representation \cite{zhang-2024-guided, salemi-etal-2024-lamp}. Related work also explores prompting or tuning LLMs with user-preference context to produce personalized outputs, highlighting the broad applicability of language-based user representations \cite{lyu-etal-2024-llm, jiang-etal-2025-reclm}.

\paragraph{Learning Personas with Downstream Objectives} A growing line of work learns user personas by directly optimizing downstream utility. Across recent preference-optimization-based approaches, the target tasks vary, but a common theme is to treat user behavior as a reward signal for persona learning. Some methods train a persona generation model that maps user histories to natural-language personas using user behavior prediction-based rewards \cite{wang2025lettingo, gao2025langptune}. Others focus on refining an existing persona over time, updating it as new behaviors arrive and optimizing updates with downstream prediction signals \cite{chen2025deeper}. Other approaches first structure long histories and then refine personas to maximize downstream utility \cite{shi2025personax, sun-etal-2025-persona}.

\section{Method}
\label{sec:method}

\paragraph{Overview.}
We propose a reward-based training framework that learns a persona model to induce high-quality user personas from intent-level memories derived from behavioral logs. Given daily behavioral logs $\mathbf{L}_d$, we use an LLM to summarize each day into intent-level memories $\mathbf{M}_d$\footnote{In our experiments, we use Gemma3-12B \cite{gemmateam2025gemma3technicalreport} for memory summarizer, with the prompt in Appendix~\ref{appendix:prompt-st}.}. Over a window $\mathbf{W}_t=[\mathbf{M}_{d_1},\ldots,\mathbf{M}_{d_t}]$, a persona model $\pi_\theta$ outputs a set of personas $\mathbf{P}_t$\footnote{The full persona-induction prompt for $\pi_\theta$ in Appendix~\ref{appendix:prompt-lt}.}:
\[
\begin{aligned}
\mathbf{L}_d &= [\ell_1,\ldots,\ell_{|\mathbf{L}_d|}] \\
\mathbf{M}_d &= \texttt{LLM}(\mathbf{L}_d) = [m_1,\ldots,m_{|\mathbf{M}_d|}] \\
\mathbf{W}_t &= [\mathbf{M}_{d_1},\ldots,\mathbf{M}_{d_t}] \\
\mathbf{P}_t &= \pi_\theta(\mathbf{W}_t) = [p_1,\ldots,p_{|\mathbf{P}_t|}]
\end{aligned}
\]
The output $\mathbf{P}_t$ is a set of tuples $p_i$ consisting of a  persona label, $K$ supporting descriptions, and an evidence set of assigned memories:
\[
\begin{aligned}
p_i = (\text{label}_i, \{\text{desc}_{i,k}\}_{k=1}^{K}, \mathcal{E}_i), \;\:
\mathcal{E}_i \subseteq \bigcup_{j=1}^{t}\mathbf{M}_{d_j}
\end{aligned}
\]
We aim to train $\pi_\theta$ so that the induced personas exhibit high representative quality with respect to their supporting memories. Specifically, we encourage (i) cohesive evidence sets, where grouped memories form semantically consistent clusters;  (ii) strong persona–evidence alignment, where the persona label and descriptions abstract the shared trait of assigned memories; and (iii) truthfulness, where persona texts do not overgeneralize or hallucinate beyond the observed evidence. In addition, we introduce global safeguards to prevent excessive omission of input memories and to discourage overly small or overly large evidence sets.

\paragraph{Candidate generation.} To construct training signals for the window $\mathbf{W}_t$, we sample $n$ candidate outputs $y^{(1)},\ldots,y^{(n)}$ from an initial policy $\pi_{0}$:
\[
\begin{aligned}
y^{(1)},\ldots,y^{(n)} \sim \pi_{0}(\cdot \mid \mathbf{W}_t)
\end{aligned}
\]
We parse each candidate $y$ into a persona set $\mathbf{P}_t$ and score it as described below.

\paragraph{Persona-quality signals.} 
We score each persona $p_i \in \mathbf{P}_t$ with a quality vector $\mathbf{S}(p_i)$:
\[ 
\begin{aligned}
\mathbf{S}(p_i) &= \big[ s_{\text{coh}}(p_i),\; s_{\text{align}}(p_i),\; s_{\text{truth}}(p_i),\; s_{\text{size}}(p_i) \big]
\end{aligned} 
\]
First, \textit{Cluster cohesion} score $s_{\text{coh}}(p_i)$ combines (i) intra-evidence cohesion with a variance penalty and (ii) separation from non-evidence memories. Here, $C_{ab}$ denotes the cosine similarity between memory embeddings, and $\bar{\mathcal{E}}_i$ denotes the memories in $\mathbf{W}_t$ not assigned to persona $p_i$:
\[
\begin{aligned}
\mu_{\text{in}} &= \mathbb{E}_{a\neq b,\, a,b\in \mathcal{E}_i}[C_{ab}] \\
\sigma_{\text{in}} &= \mathrm{Std}_{a\neq b,\, a,b\in \mathcal{E}_i}[C_{ab}] \\
\mu_{\text{cross}} &= \mathbb{E}_{a\in \mathcal{E}_i,\, b\in \bar{\mathcal{E}}_i}[C_{ab}] \\
s_{\text{coh}}(p_i) &= (\mu_{\text{in}} - \lambda \sigma_{\text{in}}) + (\mu_{\text{in}} - \mu_{\text{cross}})
\end{aligned}
\]
Second, \textit{Persona-evidence alignment} score $s_{\text{align}}$ measures how many evidence memories are judged as supporting the persona, and third, \textit{truthfulness} $s_{\text{truth}}$ evaluates whether the persona label and each description are directly supported by the evidence. Both alignment and truthfulness are computed with LLM-based judges $J_{\text{align}}$ and $J_{\text{truth}}$, prompted to output 
calibrated structured scores in $[0,1]$\footnote{Judge prompts are provided in Appendix~\ref{appendix:prompt-judge}.}:
\[
\small
\begin{aligned}
s_{\text{align}}(p_i) &= J_{\text{align}}(\text{label}_i, \mathcal{E}_i) \\
s_{\text{truth}}(p_i) &= \frac{1}{2}
\left(
J_{\text{truth}}(\text{label}_i,\mathcal{E}_i)
+ \frac{1}{K}\sum_{k=1}^{K} J_{\text{truth}}(\text{desc}_i^{k}, \mathcal{E}_i)
\right)
\end{aligned}
\]
In addition, the \textit{size score} constrains evidence-set sizes to $[e_{\min}, e_{\max}]$ and, together with cohesion, discourages overly fine- or coarse-grained clusters:
\[
\small
s_{\text{size}}(p_i) =
\begin{cases}
1-\left(\frac{e_{\min}-|\mathcal{E}_i|}{e_{\min}-1}\right)^2, & |\mathcal{E}_i| < e_{\min} \\[5pt]
1, & e_{\min}\le |\mathcal{E}_i| \le e_{\max} \\[5pt]
1-\left(\frac{|\mathcal{E}_i|-e_{\max}}{e_{\max}}\right)^2, & |\mathcal{E}_i| > e_{\max}
\end{cases}
\]
Finally, we combine these signals into a persona-level reward:
\[
\small
\begin{aligned}
r(p_i)=
\alpha_1 \cdot \frac{s_{\text{align}}(p_i) + s_{\text{truth}}(p_i) + s_{\text{coh}}(p_i)}{3}
+ \alpha_2 \cdot s_{\text{size}}(p_i)
\end{aligned}
\]

\paragraph{Scalar reward and global constraints.} To encourage the output to account for intent memories in the window $\mathbf{W}_t$, we define a coverage ratio:
\[
\small
s_{\text{cov}}(\mathbf{P}_t;\mathbf{W}_t)
=
\frac{
\left|
\bigcup_{p_i \in \mathbf{P}_t} \mathcal{E}_i
\right|
}{
\left|
\bigcup_{j=1}^{t} \mathbf{M}_{d_j}
\right|
}
\]
We apply a soft threshold:
\[
\small
\tilde s_{\text{cov}}=
\begin{cases}
1, & s_{\text{cov}} \ge 0.7,\\
s_{\text{cov}}, & \text{otherwise}.
\end{cases}
\]
We then aggregate persona-level rewards into an output-level scalar reward:
\[
\small
\begin{aligned}
r(\mathbf{P}_t;\mathbf{W}_t)= \alpha_3\,\cdot\frac{1}{|\mathbf{P}_t|}\sum_{p_i \in \mathbf{P}_t} r(p_i) + \alpha_4 \cdot \tilde s_{\text{cov}}(\mathbf{P}_t;\mathbf{W}_t)
\end{aligned}
\]
where $\alpha_1,\alpha_2,\alpha_3,\alpha_4$ are scalar weights controlling the contribution of each reward component.

\paragraph{Optimization.}
\label{subsec:method-optimization}
We transform structured persona-quality signals into a soft preference distribution within each candidate group and optimize $\pi_\theta$ to match this distribution via groupwise Direct Preference Optimization (DPO) \cite{10.5555/3666122.3668460}. For each window $\mathbf{W}_t$, we first sample $n$ candidate outputs from the initial policy $\pi_0$ and compute their scalar rewards $r^{(1)},\ldots,r^{(n)}$ as defined above.
During training, we randomly select a subset $G$:
\[
\small
\mathcal{Y}_G(\mathbf{W}_t)=\{y^{(g)}\}_{g=1}^{G} \subset \{y^{(i)}\}_{i=1}^{n}.
\]
For each selected candidate $y^{(g)}$, we get the pre-computed scalar reward:
\[
\small
r^{(g)} = r(\mathbf{P}_t^{(g)};\mathbf{W}_t),
\]
and normalize rewards within the group:
\[
\small
\tilde r^{(g)} = \frac{r^{(g)} - \mu_r}{\sigma_r + \epsilon},
\quad
q^{(g)} = \frac{\exp(\tilde r^{(g)})}{\sum_{h=1}^{G}\exp(\tilde r^{(h)})}.
\]
Let $\pi_\theta$ denote the trainable persona model and $\pi_{\mathrm{ref}}$ a frozen reference model.
For each candidate, we compute the log-probability difference:
\[
\small
\Delta^{(g)} =
\log \pi_\theta(y^{(g)} \mid \mathbf{W}_t)
-
\log \pi_{\mathrm{ref}}(y^{(g)} \mid \mathbf{W}_t).
\]
The groupwise DPO objective is then defined as:
\[
\small
\mathcal{L}
=
-\sum_{g=1}^{G}
q^{(g)}
\log
\frac{\exp(\beta \Delta^{(g)})}
{\sum_{h=1}^{G}\exp(\beta \Delta^{(h)})}
\;+\;
\lambda_{\mathrm{KL}}\cdot \widehat{\mathrm{KL}}
\]
where $\beta$ controls the sharpness of preference amplification. We use a simple deviation penalty:
\[
\small
\widehat{\mathrm{KL}}=\frac{1}{G}\sum_{g=1}^{G}\left|\Delta^{(g)}\right|.
\]
Our training objective uses graded preferences derived from reward scores while remaining fully offline over a diverse pool of generated candidates. This preserves the rich learning signal of reward-based optimization without repeated on-policy rollouts. Unlike pairwise DPO, we learn from within-group comparisons, and resampling candidate groups across updates further enriches the comparative signal.

\section{Experiment}
\subsection{Datasets}
\label{subsec:dataset}
We evaluate our method on three behavioral log datasets: one proprietary log from a large-scale online service platform (\texttt{Srv.}) and two public datasets, \texttt{MerRec}~\cite{10.1145/3690624.3709394} and \texttt{AOL}~\cite{pass2006picture}. This setup enables evaluation in both real-world and publicly reproducible settings. Table~\ref{tab:dataset_stats} summarizes key characteristics of all datasets.

\begin{table}[h!]
\small
\resizebox{\linewidth}{!}{%
\begin{tabular}{llllr}
\toprule
\textbf{Dataset} & \textbf{Type} & \textbf{Domain} & \textbf{Lang.} & \textbf{\# Interactions} \\
\midrule
\texttt{Srv.}~\footnotemark & Search / Rec. & Multiple & Korean & 19.9 ($\pm$ 12.2) \\
\texttt{MerRec} & Rec. & Shopping & English & 8.2 ($\pm$ $\;$ 9.6) \\
\texttt{AOL} & Search & Web & English & 5.2 ($\pm$ $\;$ 2.2) \\
\bottomrule
\end{tabular}%
}
\caption{Key dataset characteristics. ``Interactions'' denote user actions such as queries and clicks per user per day, computed over one-month subset from each dataset. \texttt{Srv.} exhibits about 20 interactions per user per day.}
\vspace{-0.3cm}
\label{tab:dataset_stats}
\end{table}
\footnotetext{The \texttt{Srv.} dataset covers only selected services and does not include the platform's complete logs.}

\noindent The three datasets differ in domain coverage, interaction type, and scale. \texttt{Srv.} spans diverse domains (e.g., web, news, entertainment, sports, and user-generated content) and includes search and click interactions. \texttt{MerRec} is a shopping dataset with multi-stage product interactions (view, like, add-to-cart, and purchase), while \texttt{AOL} consists of web query sequences. Example daily logs are provided in Appendix~\ref{appendix:example-log}. In scale, \texttt{Srv.} contains tens of millions of users, \texttt{MerRec} 5.56 million, and \texttt{AOL} 1,000. Based on preliminary scaling experiments on \texttt{Srv.}, we fix the training split to 500 users per dataset to control computational cost, and use 200 validation users and 300 test users for evaluation. Further analysis on the effect of training data size is reported in Appendix~\ref{appendix:data_size_exp}. 

To construct instances, we aggregate each user’s daily logs over a fixed history window $t$: two weeks for \texttt{Srv.}, which has denser logs, and four weeks for \texttt{MerRec} and \texttt{AOL}, which are sparser. For downstream evaluation, we use the items interacted with in the one-week period following $t$ as ground-truth future targets for each user in the validation and test splits. Here, items correspond to documents in \texttt{Srv.}, products in \texttt{MerRec}, and queries in \texttt{AOL}. For each dataset, we define a global candidate pool as the union of all items appearing in this future window, and evaluate how well the induced personas rank each user’s true future items. Experimental setup statistics are summarized in Table~\ref{tab:dataset_stats_exp}.

\begin{table}[h!]
\small
\resizebox{\linewidth}{!}{%
\begin{tabular}{c|clrr|cr}
\toprule
\textbf{Dataset} & \textbf{Window $t$} & \multicolumn{2}{c}{\textbf{User split}} & \multicolumn{1}{c|}{\textbf{\# Interactions}} & \textbf{Future w.} & \textbf{Items} \\ \midrule
\multirow{3}{*}{\texttt{Srv.}}   & \multirow{3}{*}{\begin{tabular}[c]{@{}c@{}}12/29$'$25 - \\ 01/11$'$26\end{tabular}} & train & 500 & 276.3 ($\pm$ 172.0)  & - & \\
                        & & val & 200 & 280.6 ($\pm$ 168.3) & \multirow{2}{*}{\begin{tabular}[c]{@{}c@{}}01/12$'$26 - \\ 01/18$'$26\end{tabular}} & \multicolumn{1}{r}{33,319} \\
                        & & test & 300 & 291.0 ($\pm$ 169.1) & & \multicolumn{1}{r}{44,414} \\ \midrule
\multirow{3}{*}{\texttt{MerRec}} & \multirow{3}{*}{\begin{tabular}[c]{@{}c@{}}05/01$'$23 - \\ 05/31$'$23\end{tabular}} & train & 500 & 106.5 ($\pm$ 233.6) & - &  \\
                        & & val & 200 & 95.1 ($\pm$ 139.0) & \multirow{2}{*}{\begin{tabular}[c]{@{}c@{}}06/01$'$23 -\\ 06/07$'$23\end{tabular}} & 2,842 \\
                        & & test & 300 & 133.7 ($\pm$ 304.6) & \multicolumn{1}{r}{} & 6,276 \\ \midrule
\multirow{3}{*}{\texttt{AOL}} & \multirow{3}{*}{\begin{tabular}[c]{@{}c@{}}03/01$'$06 -\\ 03/31$'$06\end{tabular}} & train & 500 & 120.9 ($\pm$ $\:$ 64.1) & - & \\
                        & & val & 200 & 120.9 ($\pm$ $\:$ 67.9) & \multirow{2}{*}{\begin{tabular}[c]{@{}c@{}}04/01$'$06 -\\ 04/07$'$06\end{tabular}} & 3,282 \\
                        & & test & 300 & 115.4 ($\pm$ $\:$ 62.2) & \multicolumn{1}{l}{} & 4,594 \\ \bottomrule
\end{tabular}%
}
\caption{Experimental setup statistics: time window $t$ (m/d$'$y -- m/d$'$y), train/validation/test user splits, average interactions per user within $t$, the one-week prediction window (Future w.), and the candidate item pool size.}
\label{tab:dataset_stats_exp}
\end{table}

\subsection{Models}
\paragraph{Persona Models.}
We experiment with Gemma3-27B and -12B \cite{gemmateam2025gemma3technicalreport} and Qwen3-30B and -14B \cite{yang2025qwen3technicalreport} as backbone LLMs. Each model serves as $\pi_0$ and is trained into $\pi_\theta$. This setup allows us to assess whether our framework generalizes across different model families and scales. Training hyperparameters and implementation details are provided in Appendix~\ref{appendix:exp_detail_ours}.

\paragraph{LLM-Based Reward Computation.}
For alignment and truthfulness scoring used to train $\pi_\theta$, we use Qwen3-30B as the LLM judge. This choice was informed by a preliminary qualitative assessment of its judgment quality. Further details are provided in Appendix~\ref{appendix:pre_judge_analysis}.

\paragraph{Embedding Models.}
For cohesion score computation, we use BGE-M3 \cite{chen-etal-2024-m3} for the English datasets (\texttt{MerRec}, \texttt{AOL}) and BGE-M3-ko\footnote{\url{https://huggingface.co/dragonkue/BGE-m3-ko}} for the Korean \texttt{Srv.} dataset. Qualitative examples of cluster cohesion are provided in Appendix~\ref{appendix:cohesion_score}.

\subsection{Baselines.}
We compare our method against two complementary classes of baselines: (i) frontier large-scale LLMs with high model capacity, and (ii) other clustering-based persona modeling frameworks.

\paragraph{Frontier Large-scale LLMs.} We evaluate both closed- and open-source LLMs using the same persona-induction prompt without task-specific training. Closed-source models include GPT-5.1 \cite{singh2025openaigpt5card} and Claude-4.5 \cite{claude45}. Open-source models include GPT-oss-120B \cite{openai2025gptoss120bgptoss20bmodel} and Qwen3-80B \cite{yang2025qwen3technicalreport}. This comparison examines whether improvements in persona quality can be attributed to model scale and general capacity alone.

\paragraph{Clustering-based Persona Modeling.}
PersonaX \cite{shi2025personax} constructs multiple user personas by (1) embedding user behaviors and applying hierarchical clustering to partition them, (2) selecting representative behaviors per cluster via a prototypicality-diversity trade-off, and (3) assigning persona labels to each cluster. As a clustering-based alternative to our setting, we implement PersonaX-style variants following its core design, adapted to our inputs and evaluation protocol. We evaluate two variants:
\begin{itemize}
\setlength\itemsep{0cm}
    \item \texttt{PersonaX}$_{s}$ (\textit{Summarization}), which performs embedding-based hierarchical clustering and uses an LLM to label each cluster.
    \item \texttt{PersonaX}$_{r}$ (\textit{Reflection}), which applies AgentCF \cite{10.1145/3589334.3645537} on top of the clusters to iteratively refine personas using positive and negative interaction signals.
\end{itemize}

\noindent These clustering-based baselines provide a complementary point of comparison by decoupling clustering from persona labeling and incorporating user interaction signals for persona refinement. Implementation details are provided in Appendix~\ref{appendix:exp_detail_personaX}.

\subsection{Evaluation Metrics}
We evaluate persona models along two dimensions: intrinsic persona quality and downstream utility.

\paragraph{Persona Quality.}
We report cohesion, alignment, and truthfulness scores, and their average across the three metrics as the final quality score. Each metric is computed at the persona level, averaged across personas to obtain an output-level score, and then averaged across users for final reporting. Detailed definitions and explanations of these metrics are provided in Section~\ref{sec:method}. To score alignment and truthfulness, we use the same judge as in training, Qwen3-30B, on the validation set. For the test set, we use a stronger external judge, GPT-5.1, to avoid bias from the judge used during training. 


\paragraph{Downstream Utility.}
To assess practical utility, we evaluate future interaction prediction. For each test user, we use extracted personas to rank a candidate pool of items (see Section \ref{subsec:dataset}), and examine whether they provide useful representations for predicting users' future preferences. Specifically, for each persona, we encode the persona text (i.e., its label and descriptions) and candidate items into embeddings, rank the full candidate pool using an embedding-based relevance score, compute Hit Rate at k (Hit@k) and Mean Average Precision at k (MAP@k) using the items the user interacted with as ground truth, and then average the scores across personas for each user. We report the mean user-level performance over the test set.

\subsection{Result}

\begin{table*}[t!]
\small
\resizebox{\linewidth}{!}{%
\begin{tabular}{l|llll|rrrrrr}
\toprule
 & \multicolumn{4}{c|}{Persona quality} & \multicolumn{6}{c}{Downstream utility} \\ \midrule
\multicolumn{1}{c|}{\textbf{}} & \multicolumn{1}{c|}{\textbf{Coh.}} & \multicolumn{1}{c|}{\textbf{Align.}} & \multicolumn{1}{c|}{\textbf{Truth.}} & \multicolumn{1}{c|}{\textbf{Final score}} & \multicolumn{1}{c}{\multirow{1}{*}{\textbf{H@10}}} & \multicolumn{1}{c}{\multirow{1}{*}{\textbf{H@50}}} & \multicolumn{1}{c|}{\multirow{1}{*}{\textbf{H@100}}} & \multicolumn{1}{c}{\multirow{1}{*}{\textbf{M@10}}} & \multicolumn{1}{c}{\multirow{1}{*}{\textbf{M@50}}} & \multicolumn{1}{c}{\multirow{1}{*}{\textbf{M@100}}} \\[1pt] \toprule 

\multicolumn{11}{c}{\textbf{\texttt{Srv.}}} \\ \toprule
GPT-5.1 & \multicolumn{1}{l|}{0.411 \scriptsize{($\pm$ 0.09)}} & \multicolumn{1}{l|}{0.904 \scriptsize{($\pm$ 0.07)}} & \multicolumn{1}{l|}{0.725 \scriptsize{($\pm$ 0.07)}} & 0.679 \scriptsize{($\pm$ 0.05)} & 0.2519 & 0.5067 & \multicolumn{1}{r|}{0.6387} & 0.0016 & 0.0032 & 0.0041 \\

Claude-4.5 & \multicolumn{1}{l|}{0.416 \scriptsize{($\pm$ 0.13)}} & \multicolumn{1}{l|}{\textbf{0.964} \scriptsize{($\pm$ 0.03)}} & \multicolumn{1}{l|}{\underline{0.769} \scriptsize{($\pm$ 0.07)}} & \underline{0.716} \scriptsize{($\pm$ 0.06)} & \underline{0.2765} & 0.5058 & \multicolumn{1}{r|}{0.6266} & \underline{0.0021} & \underline{0.0043} & \underline{0.0055} \\ \midrule

GPT-oss-120b & \multicolumn{1}{l|}{\underline{0.417} \scriptsize{($\pm$ 0.12)}} & \multicolumn{1}{l|}{0.926 \scriptsize{($\pm$ 0.07)}} & \multicolumn{1}{l|}{0.690 \scriptsize{($\pm$ 0.09)}} & 0.677 \scriptsize{($\pm$ 0.06)} & 0.2151 & 0.4151 & \multicolumn{1}{r|}{0.5504} & 0.0013 & 0.0028 & 0.0036 \\

Qwen3-80b & \multicolumn{1}{l|}{0.366 \scriptsize{($\pm$ 0.10)}} & \multicolumn{1}{l|}{0.805 \scriptsize{($\pm$ 0.12)}} & \multicolumn{1}{l|}{0.643 \scriptsize{($\pm$ 0.08)}} & 0.604 \scriptsize{($\pm$ 0.06)} & 0.2566 & \underline{0.5091} & \multicolumn{1}{r|}{\underline{0.6431}} & 0.0017 & 0.0036 & 0.0047 \\ \midrule

$\text{PersonaX}_s$ & \multicolumn{1}{l|}{0.349 \scriptsize{($\pm$ 0.07)}} & \multicolumn{1}{l|}{0.913 \scriptsize{($\pm$ 0.07)}} & \multicolumn{1}{l|}{0.707 \scriptsize{($\pm$ 0.06)}} & 0.656 \scriptsize{($\pm$ 0.05)} & 0.2044 & 0.4362 & \multicolumn{1}{r|}{0.5806} & 0.0012 & 0.0024 & 0.0031 \\

$\text{PersonaX}_{r}$ & \multicolumn{1}{l|}{0.345 \scriptsize{($\pm$ 0.07)}} & \multicolumn{1}{l|}{0.656 \scriptsize{($\pm$ 0.15)}} & \multicolumn{1}{l|}{0.507 \scriptsize{($\pm$ 0.07)}} & 0.502 \scriptsize{($\pm$ 0.07)} & 0.1346 & 0.3546 & \multicolumn{1}{r|}{0.4929} & 0.0004 & 0.0008 & 0.0011 \\ \midrule

$\pi_\theta$ (Ours) & \multicolumn{1}{l|}{\textbf{0.455} \scriptsize{($\pm$ 0.08)}} & \multicolumn{1}{l|}{\underline{0.961} \scriptsize{($\pm$ 0.04)}} & \multicolumn{1}{l|}{\textbf{0.893} \scriptsize{($\pm$ 0.03)}} & \textbf{0.769} \scriptsize{($\pm$ 0.03)} & \textbf{0.3139} & \textbf{0.6071} & \multicolumn{1}{r|}{\textbf{0.7370}} & \textbf{0.0025} & \textbf{0.0056} & \textbf{0.0075} \\[1pt] \toprule

\multicolumn{11}{c}{\textbf{\texttt{MerRec}}} \\ \toprule
GPT-5.1 & \multicolumn{1}{l|}{0.553 \scriptsize{($\pm$ 0.20)}} & \multicolumn{1}{l|}{0.925 \scriptsize{($\pm$ 0.07)}} & \multicolumn{1}{l|}{0.612 \scriptsize{($\pm$ 0.09)}} & 0.696 \scriptsize{($\pm$ 0.08)} & 0.3984 & 0.5581 & \multicolumn{1}{r|}{0.6315} & 0.0455 & 0.0642 & 0.0718  \\

Claude-4.5 & \multicolumn{1}{l|}{0.523 \scriptsize{($\pm$ 0.24)}} & \multicolumn{1}{l|}{\textbf{0.959} \scriptsize{($\pm$ 0.05)}} & \multicolumn{1}{l|}{0.624 \scriptsize{($\pm$ 0.11)}} & 0.701 \scriptsize{($\pm$ 0.10)} & 0.4275 & 0.5693 & \multicolumn{1}{r|}{0.6307} & 0.0487 & 0.0717 & 0.0808 \\ \midrule

GPT-oss-120b & \multicolumn{1}{l|}{0.489 \scriptsize{($\pm$ 0.24)}} & \multicolumn{1}{l|}{\underline{0.945} \scriptsize{($\pm$ 0.07)}} & \multicolumn{1}{l|}{0.599 \scriptsize{($\pm$ 0.11)}} & 0.677 \scriptsize{($\pm$ 0.10)} & 0.3525 & 0.5004 & \multicolumn{1}{r|}{0.5916} &  0.0391 & 0.0554 & 0.0618 \\

Qwen-80b & \multicolumn{1}{l|}{0.496 \scriptsize{($\pm$ 0.21)}} & \multicolumn{1}{l|}{0.883 \scriptsize{($\pm$ 0.12)}} & \multicolumn{1}{l|}{0.547 \scriptsize{($\pm$ 0.11)}} & 0.641 \scriptsize{($\pm$ 0.08)} & 0.4259 & 0.5723 & \multicolumn{1}{r|}{0.6424} & \underline{0.0492} & 0.0712 & 0.0790 \\ \midrule

$\text{PersonaX}_s$ & \multicolumn{1}{l|}{\underline{0.557} \scriptsize{($\pm$ 0.12)}} & \multicolumn{1}{l|}{0.915 \scriptsize{($\pm$ 0.09)}} & \multicolumn{1}{l|}{\underline{0.635} \scriptsize{($\pm$ 0.11)}} & \underline{0.702} \scriptsize{($\pm$ 0.07)} & \underline{0.5334} & \underline{0.6935} & \multicolumn{1}{r|}{\underline{0.7523}} & 0.0440 & \underline{0.0752} & \underline{0.0877} \\

$\text{PersonaX}_{r}$ & \multicolumn{1}{l|}{0.555 \scriptsize{($\pm$ 0.12)}} & \multicolumn{1}{l|}{0.768 \scriptsize{($\pm$ 0.19)}} & \multicolumn{1}{l|}{0.417 \scriptsize{($\pm$ 0.10)}} & 0.580 \scriptsize{($\pm$ 0.10)} & 0.2938 & 0.4931 & \multicolumn{1}{r|}{0.5933} & 0.0115 & 0.0207 & 0.0250 \\ \midrule

$\pi_\theta$ (Ours) & \multicolumn{1}{l|}{\textbf{0.674} \scriptsize{($\pm$ 0.17)}} & \multicolumn{1}{l|}{0.894 \scriptsize{($\pm$ 0.10)}} & \multicolumn{1}{l|}{\textbf{0.798} \scriptsize{($\pm$ 0.09)}} & \textbf{0.788} \scriptsize{($\pm$ 0.09)} & \textbf{0.5865} & \textbf{0.7135} & \multicolumn{1}{r|}{\textbf{0.7692}} & \textbf{0.1065} & \textbf{0.1432} & \textbf{0.1557} \\[1pt] \toprule 

\multicolumn{11}{c}{\textbf{\texttt{AOL}}} \\ \toprule
GPT-5.1 & \multicolumn{1}{l|}{0.605 \scriptsize{($\pm$ 0.13)}} & \multicolumn{1}{l|}{0.942 \scriptsize{($\pm$ 0.05)}} & \multicolumn{1}{l|}{0.607 \scriptsize{($\pm$ 0.07)}} & 0.718 \scriptsize{($\pm$ 0.05)} & 0.2927 & 0.4879 & \multicolumn{1}{r|}{\underline{0.5900}} & 0.0265 & 0.0316 & 0.0329 \\

Claude-4.5 & \multicolumn{1}{l|}{0.563 \scriptsize{($\pm$ 0.15)}} & \multicolumn{1}{l|}{\textbf{0.959} \scriptsize{($\pm$ 0.05)}} & \multicolumn{1}{l|}{0.635 \scriptsize{($\pm$ 0.08)}} & 0.718 \scriptsize{($\pm$ 0.07)} & 0.2601 & 0.4474 & \multicolumn{1}{r|}{0.5714} & 0.0216 & 0.0260 & 0.0271 \\ \midrule

GPT-oss-120b & \multicolumn{1}{l|}{\underline{0.610} \scriptsize{($\pm$ 0.17)}} & \multicolumn{1}{l|}{\underline{0.955} \scriptsize{($\pm$ 0.07)}} & \multicolumn{1}{l|}{\underline{0.654} \scriptsize{($\pm$ 0.08)}} & \underline{0.739} \scriptsize{($\pm$ 0.07)} & 0.2695 & 0.4355 & \multicolumn{1}{r|}{0.5533} & 0.0241 & 0.0286 & 0.0297 \\

Qwen3-80b & \multicolumn{1}{l|}{0.573 \scriptsize{($\pm$ 0.13)}} & \multicolumn{1}{l|}{0.835 \scriptsize{($\pm$ 0.11)}} & \multicolumn{1}{l|}{0.546 \scriptsize{($\pm$ 0.08)}} & 0.651 \scriptsize{($\pm$ 0.06)} & 0.3055 & 0.4689 & \multicolumn{1}{r|}{0.5850} & \underline{0.0281} & \underline{0.0334} & \underline{0.0347} \\ \midrule

$\text{PersonaX}_s$ & \multicolumn{1}{l|}{0.597 \scriptsize{($\pm$ 0.09)}} & \multicolumn{1}{l|}{0.865 \scriptsize{($\pm$ 0.09)}} & \multicolumn{1}{l|}{0.599 \scriptsize{($\pm$ 0.08)}} & 0.686 \scriptsize{($\pm$ 0.06)} & \underline{0.3229} & \underline{0.4898} & \multicolumn{1}{r|}{0.5867} & 0.0277 & 0.0317 & 0.0329 \\

$\text{PersonaX}_{r}$ & \multicolumn{1}{l|}{0.594 \scriptsize{($\pm$ 0.09)}} & \multicolumn{1}{l|}{0.715 \scriptsize{($\pm$ 0.15)}} & \multicolumn{1}{l|}{0.428 \scriptsize{($\pm$ 0.08)}} & 0.578 \scriptsize{($\pm$ 0.08)} & 0.1211 & 0.3259 & \multicolumn{1}{r|}{0.4620} & 0.0042 & 0.0059 & 0.0065 \\ \midrule

$\pi_\theta$ (Ours) & \multicolumn{1}{l|}{\textbf{0.718} \scriptsize{($\pm$ 0.12)}} & \multicolumn{1}{l|}{0.944 \scriptsize{($\pm$ 0.05)}} & \multicolumn{1}{l|}{\textbf{0.791} \scriptsize{($\pm$ 0.06)}} & \textbf{0.817} \scriptsize{($\pm$ 0.05)} & \textbf{0.4018} & \textbf{0.5817} & \multicolumn{1}{r|}{\textbf{0.6782}} & \textbf{0.0445} & \textbf{0.0517} & \textbf{0.0531} \\ 

\bottomrule
\end{tabular}%
}
\caption{Main results on the test sets of \texttt{Srv.}, \texttt{MerRec}, and \texttt{AOL}. We compare frontier closed- and open-source LLMs in a prompt-only setting, along with our model ($\pi_\theta$) and PersonaX-style baselines built on Gemma3-27B, on both persona quality and downstream utility. Persona quality is evaluated using cluster cohesion (Coh.), persona--evidence alignment (Align.), persona truthfulness (Truth.), and their average as the final score. Downstream utility is measured by future interaction prediction using Hit@k and MAP@k. For each dataset and metric, the best result is shown in \textbf{bold} and the second-best result is \underline{underlined}.}
\vspace{-0.3cm}
\label{tab:result_main}
\end{table*}

Table~\ref{tab:result_main} presents the main results on three datasets. Across all datasets, our model outperforms both frontier LLMs and PersonaX-style baselines in final persona quality, and this advantage transfers consistently to downstream prediction, where it achieves the best Hit@k and MAP@k. Although some frontier LLMs obtain comparable or higher alignment scores, likely due to their strength in generating fluent persona descriptions that explain memories well at the textual level, our method produces more cohesive clusters and more truthful personas through explicit persona-quality optimization. Unlike PersonaX$_s$, which summarizes pre-formed clusters one at a time, and PersonaX$_r$, which refines personas by iteratively processing cluster memories with external negative signals, our method jointly optimizes grouping and abstraction over the full memory context, enabling the induction of more distinct personas that better capture shared behavioral themes.

\vspace{-0.1cm}
\begin{table}[h!]
\centering
\small
\resizebox{0.9\linewidth}{!}{%
\begin{tabular}{lc|r|r|r|r}
\toprule
 & \multicolumn{1}{l|}{} & \multicolumn{2}{c|}{\textbf{\texttt{Srv.}}} & \multicolumn{2}{c}{\textbf{\texttt{MerRec}}} \\[1pt]
                           & \multicolumn{1}{l|}{} & \multicolumn{1}{c|}{\small{Quality}} & \multicolumn{1}{c|}{\small{H@100}} & \multicolumn{1}{c|}{\small{Quality}} & \multicolumn{1}{c}{\small{H@100}} \\ \midrule

\multirow{2}{*}{Qwen3-14B} & $\pi_0$ & 0.68 & 0.55 & 0.72 & 0.61 \\
                           & $\pi_\theta$ & 0.73 & 0.64 & 0.75 & 0.63 \\ \midrule

\multirow{2}{*}{Qwen3-30B} & $\pi_0$ & 0.64 & 0.58 & 0.65 & 0.60 \\
                           & $\pi_\theta$ & 0.72 & 0.68 & 0.75 & 0.70 \\ \midrule

\multirow{2}{*}{Gemma3-12B} & $\pi_0$ & 0.61 & 0.53 & 0.68 & 0.60 \\
                            & $\pi_\theta$ & 0.68 & 0.67 & 0.76 & 0.69 \\ \midrule

\multirow{2}{*}{Gemma3-27B} & $\pi_0$ & 0.67 & 0.50 & 0.69 & 0.57 \\
                            & $\pi_\theta$ & 0.77 & 0.74 & 0.79 & 0.77 \\ \bottomrule
\end{tabular}%
}
\caption{Generalization across different models on the test set. We train multiple backbone LLMs and evaluate performance before ($\pi_0$) and after training ($\pi_\theta$). The best checkpoint for each model is selected based on final persona quality score on the validation set. Here, Quality denotes the average of Coh., Align., and Truth.}
\vspace{-0.5cm}
\label{tab:result_generalize}
\end{table}



\begin{figure*}[t!]
    \centering
    \includegraphics[width=1.0\linewidth]{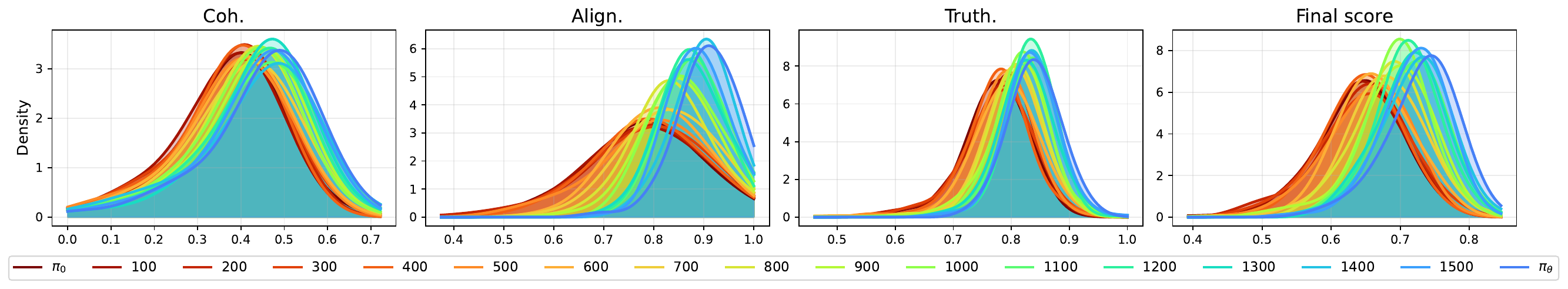}
    \caption{Score distributions on the validation set during training for our method on \texttt{Srv.} dataset. Each curve shows the distribution of persona-level cohesion, alignment, truthfulness, and final scores at different training steps, from the initial model ($\pi_0$) to the selected model ($\pi_\theta$, step 1600).}
    \label{fig:reward_distribution}
    \vspace{-0.3cm}
\end{figure*}
\begin{figure}[t!]
    \centering
    \includegraphics[width=1.0\linewidth]{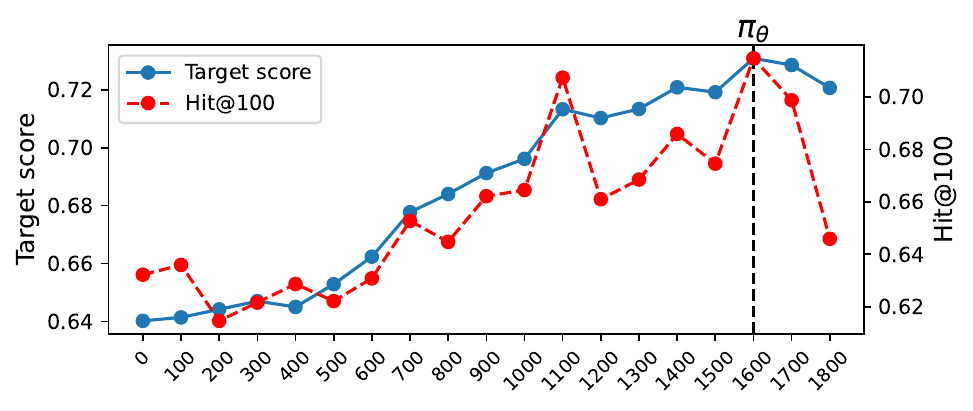}
    \caption{Validation performance during training on \texttt{Srv.} dataset. We plot the target score, defined as the average of cohesion, alignment, and truthfulness, together with downstream utility (Hit@100) across steps.}
    \label{fig:val_quality_hit}
    \vspace{-0.2cm}
\end{figure}

Table~\ref{tab:result_generalize} evaluates the generalizability of our method across different backbone LLMs. Across models, training consistently improves both persona quality and downstream performance over the initial $\pi_0$, indicating that our method is not tied to a specific backbone.  Furthermore, Figure~\ref{fig:reward_distribution} shows how the distributions of persona quality scores on the validation set evolve during training. Starting from the initial model $\pi_0$, the distributions progressively shift toward higher scores, indicating improvements in cohesion, alignment, and truthfulness as optimization proceeds. Finally, Figure~\ref{fig:val_quality_hit} illustrates the relationship between persona quality and downstream performance. As training progresses, the target score increases, while the downstream metric (Hit@100) improves in parallel. This suggests that optimizing intrinsic persona quality can also lead to better downstream utility.

\section{Analysis}
In this section, we further analyze our hierarchical persona-induction framework. All analyses are conducted using our model trained from Gemma3-27B. We focus on three aspects of the proposed approach: (1) how behavioral logs are progressively compressed into higher-level representations, (2) the contribution of each reward component through ablation, and (3) qualitative differences between personas produced before and after training.

\subsection{Log-to-Persona Compression}
Our framework constructs hierarchical abstractions of user behavior by progressively compressing raw behavioral logs into intent memories and finally into a small set of personas. Table~\ref{tab:analysis_size} shows the average number of units at each representation level within the time window $t$. Across datasets, the hierarchy dramatically reduces the number of units. For example, on \texttt{Srv.}, an average of 291 raw log entries per user are summarized into 83.6 intent memories and further distilled into only 4.8 personas. Similar compression patterns appear in \texttt{MerRec} and \texttt{AOL}, indicating that the hierarchical aggregation effectively condenses long behavioral histories into compact representations. Importantly, the final persona layer remains highly predictive despite using far fewer units than intent memories. Across datasets, persona-level representations achieve comparable or stronger Hit@100, suggesting that the final abstraction captures salient aspects of user behavior.

\begin{table}[h!]
\centering
\small
\resizebox{0.8\linewidth}{!}{%
\begin{tabular}{llrrr}
\toprule
\multicolumn{1}{l}{}     &         & \multicolumn{1}{c}{\texttt{\textbf{Srv.}}} & \multicolumn{1}{c}{\texttt{\textbf{MerRec}}} & \multicolumn{1}{c}{\texttt{\textbf{AOL}}} \\ \midrule
\textbf{Log} & \scriptsize{Count} & 291.0 & 133.7 & 115.4 \\ \midrule
\multirow{2}{*}{\textbf{Memory}}  & \scriptsize{Count}   & 83.6 & 42.6 & 65.3 \\
                         & \scriptsize{Hit@100} & 0.66 & 0.79 & 0.64 \\ \midrule
\multirow{2}{*}{\textbf{Persona}} & \scriptsize{Count}   & 4.8 & 2.4 & 5.1 \\
                         & \scriptsize{Hit@100} & 0.74 & 0.77 & 0.68 \\ \bottomrule
\end{tabular}%
}
\caption{
Average number of units per user (Count) and future prediction performance (Hit@100) across representation levels within window $t$: raw logs, intent memories, and personas.
}
\vspace{-0.3cm}
\label{tab:analysis_size}
\end{table}

\subsection{Reward Ablation}
To better understand the contribution of each reward component, we conduct an ablation study by removing one reward at a time during training. Table~\ref{tab:analysis_abl} reports the resulting persona quality scores and downstream prediction performance.

\begin{table}[h!]
\centering
\small
\resizebox{0.9\linewidth}{!}{%
\begin{tabular}{l|rrrr|rr}
\toprule
         & \textbf{Coh.} & \textbf{Align.} & \textbf{Truth.} & \multicolumn{1}{c|}{\textbf{Final}} & \multicolumn{1}{c}{\textbf{H@100}} \\ \midrule
Full      & \textit{0.455} & 0.961 & \textbf{0.893} & \textbf{0.769} & \textbf{0.737} \\
\scriptsize{-- Align.} & 0.457 & \textit{0.929} & 0.876 & \textit{0.754} & 0.724 \\
\scriptsize{-- Coh.}   & 0.469 & \textbf{0.967} & 0.861 & 0.765 & \textit{0.688} \\
\scriptsize{-- Truth.} & \textbf{0.483} & 0.965 & \textit{0.840} & 0.762 & 0.726 \\ \bottomrule
\end{tabular}%
}
\caption{Reward ablation results on the \texttt{Srv.} dataset. Each row removes one reward component during training. The best checkpoint is selected based on validation set persona quality and evaluated on the test set. For each metric, the best result is shown in \textbf{bold} and the worst result in \textit{italic}.}
\vspace{-0.1cm}
\label{tab:analysis_abl}
\end{table}


\noindent Table~\ref{tab:analysis_abl} suggests that the reward components contribute in different ways. Removing the alignment reward lowers both the alignment score and H@100, indicating that this signal helps tie personas to their supporting memories in a way that also benefits downstream prediction. The truthfulness reward behaves more independently, mainly affecting the truthfulness score while leaving the other metrics relatively stable. Most interestingly, removing the cohesion reward does not reduce the measured cohesion score and leaves the final quality score nearly unchanged, yet it causes the largest drop in H@100. This may suggest that the measured cohesion score is partly entangled with alignment, while the explicit embedding-based cohesion reward provides an additional structural signal during training that helps the model distinguish among candidate clusters and favor ones that better reflect underlying behavioral patterns.

\subsection{Qualitative Comparison of $\pi_0$ and $\pi_\theta$}
\begin{figure}[htp!]
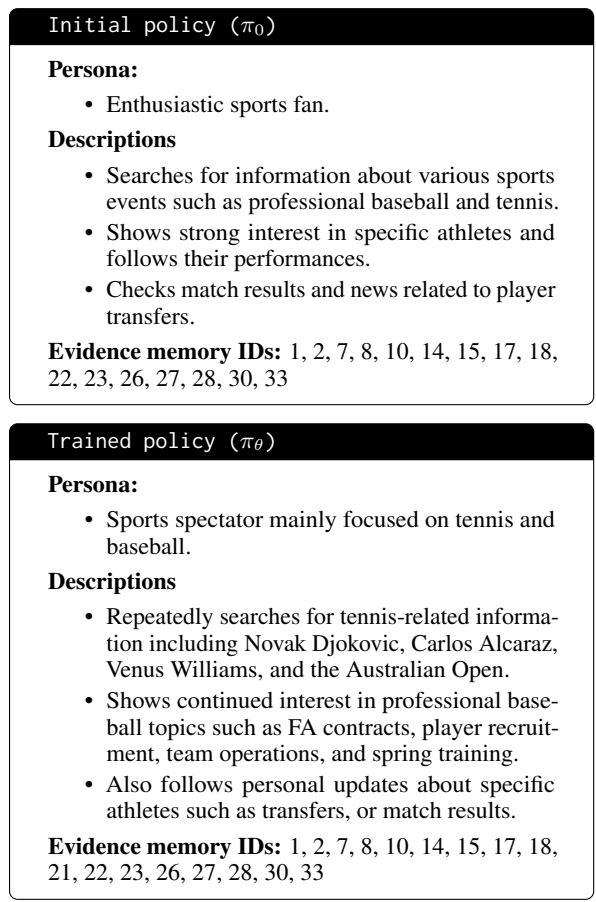

\centering
\begin{tcolorbox}[colback=white,colframe=black,title=\texttt{\small{Initial policy ($\pi_0$)}}]
\small
\vspace{-0.1cm}
\textbf{Persona:}
\vspace{-0.2cm}
\begin{itemize}
\setlength\itemsep{-0.1cm}
\item Enthusiastic sports fan.
\end{itemize}

\vspace{-0.2cm}
\textbf{Descriptions}
\vspace{-0.2cm}
\begin{itemize}
\setlength\itemsep{-0.1cm}
\item Searches for information about various sports events such as professional baseball and tennis.
\item Shows strong interest in specific athletes and follows their performances.
\item Checks match results and news related to player transfers.
\end{itemize}

\vspace{-0.2cm}
\textbf{Evidence memory IDs:} 1, 2, 7, 8, 10, 14, 15, 17, 18, 22, 23, 26, 27, 28, 30, 33
\vspace{-0.1cm}
\end{tcolorbox}

\begin{tcolorbox}[colback=white,colframe=black,title=\texttt{\small{Trained policy ($\pi_\theta$)}}]
\small
\vspace{-0.1cm}
\textbf{Persona:}
\vspace{-0.2cm}
\begin{itemize}
\setlength\itemsep{-0.1cm}
\item Sports spectator mainly focused on tennis and baseball.
\end{itemize}

\vspace{-0.2cm}
\textbf{Descriptions}
\vspace{-0.2cm}
\begin{itemize}
\setlength\itemsep{-0.1cm}
\item Repeatedly searches for tennis-related information including Novak Djokovic, Carlos Alcaraz, Venus Williams, and the Australian Open.
\item Shows continued interest in professional baseball topics such as FA contracts, player recruitment, team operations, and spring training.
\item Also follows personal updates about specific athletes such as transfers, or match results.
\end{itemize}

\vspace{-0.2cm}
\textbf{Evidence memory IDs:} 1, 2, 7, 8, 10, 14, 15, 17, 18, 21, 22, 23, 26, 27, 28, 30, 33
\vspace{-0.1cm}
\end{tcolorbox}
\caption{Generated personas before and after training with largely overlapping evidence memories.}
\vspace{-0.1cm}
\label{fig:persona_example}
\end{figure}

To illustrate how training improves persona induction, we compare personas generated by the initial policy $\pi_0$ and the trained policy $\pi_\theta$ from the same input on two representative examples from the \texttt{Srv.} test set. These examples highlight two common forms of improvement. In some cases, the trained model expresses essentially the same evidence cluster with a clearer characterization of the user. In other cases, it refines the granularity of the clustering itself, yielding a more coherent cluster together with a sharper persona label. First, as illustrated in Figure~\ref{fig:persona_example}, the trained model identifies a more specific sports-viewing pattern centered on tennis and baseball, compared to the broad role-based summary produced by $\pi_0$.

\begin{figure}[htp!]
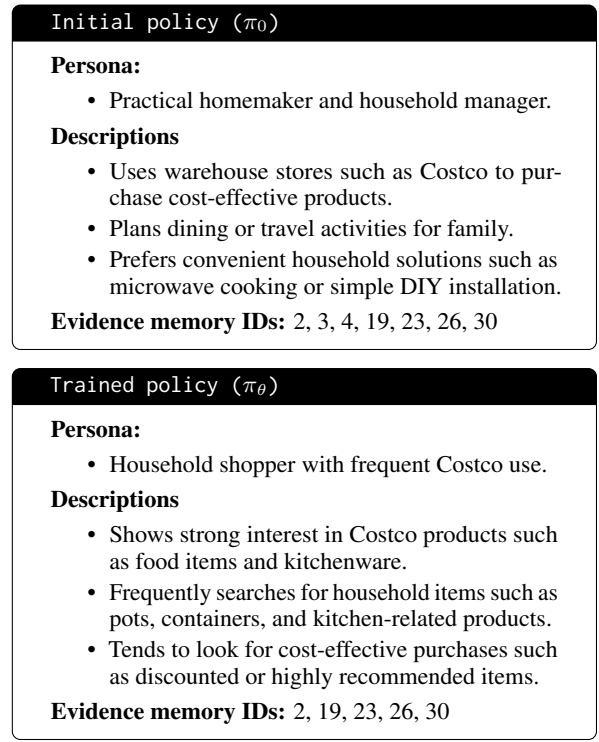

\centering
\begin{tcolorbox}[breakable=false, colback=white,colframe=black,title=\texttt{\small{Initial policy ($\pi_0$)}}]
\small
\vspace{-0.1cm}
\textbf{Persona:}
\vspace{-0.2cm}
\begin{itemize}
\setlength\itemsep{-0.1cm}
\item Practical homemaker and household manager.
\end{itemize}

\vspace{-0.2cm}
\textbf{Descriptions}
\vspace{-0.2cm}
\begin{itemize}
\setlength\itemsep{-0.1cm}
\item Uses warehouse stores such as Costco to purchase cost-effective products.
\item Plans dining or travel activities for family.
\item Prefers convenient household solutions such as microwave cooking or simple DIY installation.
\end{itemize}

\vspace{-0.2cm}
\textbf{Evidence memory IDs:} 2, 3, 4, 19, 23, 26, 30
\vspace{-0.1cm}
\end{tcolorbox}

\begin{tcolorbox}[breakable=false,colback=white,colframe=black,title=\texttt{\small{Trained policy ($\pi_\theta$)}}]
\small
\vspace{-0.1cm}
\textbf{Persona:}
\vspace{-0.2cm}
\begin{itemize}
\setlength\itemsep{-0.1cm}
\item Household shopper with frequent Costco use.
\end{itemize}

\vspace{-0.2cm}
\textbf{Descriptions}
\vspace{-0.2cm}
\begin{itemize}
\setlength\itemsep{-0.1cm}
\item Shows strong interest in Costco products such as food items and kitchenware.
\item Frequently searches for household items such as pots, containers, and kitchen-related products.
\item Tends to look for cost-effective purchases such as discounted or highly recommended items.
\end{itemize}

\vspace{-0.2cm}
\textbf{Evidence memory IDs:} 2, 19, 23, 26, 30
\vspace{-0.1cm}
\end{tcolorbox}
\caption{Generated personas before and after training showing clearer abstraction granularity.}
\vspace{-0.2cm}
\label{fig:persona_example_2}
\end{figure}

\noindent A different type of improvement is illustrated in Figure~\ref{fig:persona_example_2}. While $\pi_0$ summarizes the evidence with a broad household role, the trained model produces a more coherent cluster centered on a dominant behavioral pattern—frequent Costco-related shopping—and assigns a more precise persona label. Additional qualitative examples from \texttt{MerRec} and \texttt{AOL} datasets are provided in Appendix~\ref{appendix:example-lt}.

\section{Conclusion}
We presented a hierarchical framework for inducing evidence-grounded natural-language personas from behavioral logs. By explicitly defining persona quality in terms of cohesion, alignment, and truthfulness, and optimizing these signals with an offline RL-style objective, our method produces personas that are both high-quality and useful for downstream prediction. Our analyses further show that the hierarchy compresses long behavioral histories into compact yet informative representations, and that the reward components play complementary roles in training. Taken together, our results suggest that persona quality is closely tied to downstream utility, as both reflect whether personas can serve as reliable and useful user representations. Limitations and practical considerations of the proposed framework are discussed in Section~\ref{sec:limitation}.

\section{Limitations}
\label{sec:limitation}
While our framework shows strong promise across datasets and models, several limitations and practical considerations remain.

\paragraph{LLM-based evaluation.}
Our evaluation of alignment and truthfulness relies on LLM-based judges. While this allows scalable evaluation across large datasets, the scores may still depend on the behavior of the judging model. To mitigate this, we use a stronger external model for test set evaluation, but exploring alternative evaluation protocols or multiple judges could further improve robustness.

\paragraph{Training paradigm.}
Our persona model is trained using offline reinforcement learning based on pre-generated candidate outputs. This makes experimentation more practical by avoiding repeated on-policy rollouts, but other training paradigms remain underexplored. In particular, online RL approaches such as GRPO-style updates \cite{shao2024deepseekmathpushinglimitsmathematical, guo2025deepseek} could be explored to further improve persona induction.

\paragraph{Temporal persona management.}
Our framework induces personas independently within a fixed behavioral window $t$. In practice, however, user behavior evolves continuously over time, and systems may accumulate personas across many windows. A remaining practical challenge is how newly induced personas should update an existing persona bank, such as preserving stable personas, refining existing ones, or merging overlapping personas as user behavior evolves. Developing an effective persona bank update framework is an important direction for our future work.

\paragraph{Limited downstream utility evaluation.} Although induced natural-language personas may be useful for a range of personalized downstream applications, in this work we evaluate their utility on a single downstream task: future interaction prediction based on relevance ranking. While this is a natural and practically important setting, as personas can serve as user-side representations for ranking future items, exploring their use in broader user interaction systems and across a wider range of downstream tasks remains important future work.

\paragraph{Potential Risks.}
Our framework induces personas from behavioral logs and may therefore raise concerns about user profiling or unintended use of inferred user characteristics. These risks are particularly important in applications involving sensitive user data, and should be taken into account when considering real-world deployment.

\bibliography{custom}

\appendix
\clearpage
\section{Appendix}

\subsection{Additional Analysis}
\subsubsection{Qualitative Examples from the LLM Judge}
\label{appendix:pre_judge_analysis}
As a preliminary qualitative check, we inspected the judgments produced by Qwen3-30B on a small set of persona--evidence examples sampled from the training data. We found that it generally produced stable and reliable scores, penalizing unsupported persona claims while rewarding abstractions that were well grounded in the evidence memories. Based on this observation, we use Qwen3-30B as the LLM judge to assign rewards to offline candidates generated by the initial policy $\pi_0$ before training. Below, we provide example judgments for both alignment and truthfulness from the \texttt{Srv.} dataset. Each example shows the generated persona together with selected evidence memories.

\paragraph{Persona-Evidence Alignment} $\:$ In the high-scoring example below, the persona closely reflects the dominant topic shared by the evidence memories. The memories consistently concern sports players, injuries, trades, and team roster changes, and the persona summarizes this theme by describing systematic interest in team tactics and player analysis. In contrast, the low-scoring example shows a persona about household storage planning, while most evidence memories concern travel planning and accommodation searches. Because the persona does not reflect the main topics present in the evidence memories, the judge assigns a substantially lower alignment score.
\vspace{0.1cm}

\begin{tcolorbox}[colback=white,colframe=black,title=\texttt{\small{High alignment (score = 1.0)}}]
\small
\textbf{Persona:} 
\vspace{-0.2cm}
\begin{itemize}
\setlength\itemsep{-0.1cm}
\item Systematic interest in tactical changes involving sports teams and players.
\end{itemize}

\vspace{-0.1cm}
\textbf{Descriptions}
\vspace{-0.2cm}
\begin{itemize}
\setlength\itemsep{-0.1cm}
\item Analyzes team performance and tactics by focusing on players’ individual skills, injury history, and trade news.
\item Evaluates players’ performance and roles through comparative analysis between athletes.
\item Attempts to understand team strategy by connecting game results with roster changes and tactical adjustments.
\end{itemize}

\vspace{-0.1cm}
\textbf{Evidence memory IDs:} 0, 3, 4, 5, 8, 10, 21, 22, 26, 27
\vspace{-0.2cm}
\begin{itemize}
\setlength\itemsep{-0.1cm}
\item \textbf{0.} Shai Gilgeous-Alexander performance and injury updates: Search activity about Shai Gilgeous-Alexander’s game performance, scoring results, and injury status indicates interest in player health and match outcomes.
\item \textbf{3.} Anthony Davis injuries and trade discussions: Search logs about Anthony Davis’ recurring injuries and potential trade scenarios suggest interest in how player conditions affect team performance.
\item \textbf{4.} Gangwon FC signing Ko Young-jun: Search activity about Gangwon FC recruiting Ko Young-jun, an Asian Games gold medalist, indicates interest in team roster reinforcement.
\item \textbf{5.} KIA free-agent negotiation involving Cho Sang-woo: Search logs about free-agent negotiations and contract issues involving Cho Sang-woo reflect interest in player transfers and contract discussions.
\item \textbf{8.} Comparison between Giannis Antetokounmpo and LeBron James: Search activity comparing the achievements and performance of Giannis Antetokounmpo and LeBron James indicates interest in evaluating elite players.
\item \textbf{10.} Analysis of Trae Young and Luka Dončić: Search logs about Trae Young and Luka Dončić analyzing their playing styles and potential trade discussions indicate interest in player performance and team tactics.
\item \textbf{21.} Boston Celtics roster and trade analysis: Search activity about Boston Celtics player trades and roster restructuring indicates interest in tactical changes within the team.
\item \textbf{22.} Comparison between Jayson Tatum and Jaylen Brown: Search logs comparing the performance and roles of Jayson Tatum and Jaylen Brown suggest interest in evaluating key players within a team.
\item \textbf{26.} Deandre Ayton performance and Lakers center recruitment: Search activity about Deandre Ayton’s performance and the Los Angeles Lakers’ search for a center reflects interest in roster strategy and player roles.
\item \textbf{27.} Game analysis of LeBron James and Luka Dončić: Search logs analyzing performances and match results involving LeBron James and Luka Dončić indicate sustained interest in NBA player performance.
\end{itemize}
\end{tcolorbox}

\begin{tcolorbox}[colback=white,colframe=black,title=\texttt{\small{Low alignment (score = 0.14)}}]
\small
\textbf{Persona:} 
\vspace{-0.2cm}
\begin{itemize}
\setlength\itemsep{-0.1cm}
\item Practical planner focused on optimizing household storage and space organization.
\end{itemize}
\vspace{-0.1cm}
\textbf{Descriptions}
\vspace{-0.2cm}
\begin{itemize}
\setlength\itemsep{-0.1cm}
\item Compares storage products for spaces such as utility rooms and children's rooms using products from retail stores.
\item Repeatedly considers factors such as price, size, durability, and ease of installation when selecting storage solutions.
\item Seeks practical and budget-friendly storage solutions to efficiently organize household space.
\end{itemize}
\vspace{-0.1cm}
\textbf{Evidence memory IDs:} 42, 57, 58, 59, 66, 67, 72
\vspace{-0.2cm}
\begin{itemize}
\setlength\itemsep{-0.1cm}
\item \textbf{42.} Utility room organization and storage product comparison: Search activity related to organizing a utility room and purchasing storage products, comparing items from lifestyle retailers to find practical storage solutions.
\item \textbf{57.} Airport transportation and travel cost information for a Southeast Asian destination: Search logs indicate interest in taxi fares, airport pickup services, lounge access, and currency exchange when traveling.
\item \textbf{58.} City sightseeing and tourism activities: Search activity about city tours, theme parks, and island hopping tours suggests exploration of tourist attractions and travel activities.
\item \textbf{59.} Family accommodation booking during travel: Search logs indicate interest in booking family-friendly accommodation, particularly hotels suitable for traveling with a young child.
\item \textbf{66.} Resort facilities and travel itinerary planning: Search activity about resort amenities, swimming pools, restaurants, and late check-out options during travel.
\item \textbf{67.} Resort accommodation options for families:  
Search logs about family rooms and hotel room types suitable for families traveling together.
\item \textbf{72.} Comparison of family hotels and room options: Search activity comparing hotel reviews, prices, and family suite options when planning travel accommodation.
\end{itemize}
\end{tcolorbox}

\vspace{0.1cm}
\paragraph{Persona Truthfulness} $\:$ In the high-scoring example below, the persona and its descriptions are directly supported by the underlying evidence memories, which consistently concern health conditions, treatments, and related management practices. The persona therefore accurately summarizes the shared theme of the evidence without introducing unsupported claims. In contrast, the low-scoring example shows a persona describing a self-improvement routine and productivity-oriented exploration, while the evidence memories mainly concern university admissions. Because the persona introduces behavioral patterns that are not grounded in the evidence, the judge assigns a lower truthfulness score.
\vspace{0.1cm}

\begin{tcolorbox}[colback=white,colframe=black,title=\texttt{\small{High truthfulness (score = 1.0)}}]
\small
\textbf{Persona:} 
\vspace{-0.2cm}
\begin{itemize}
\setlength\itemsep{-0.1cm}
\item Health-conscious individual who repeatedly searches for specific management information for health issues.
\end{itemize}
\vspace{-0.1cm}
\textbf{Descriptions}
\vspace{-0.2cm}
\begin{itemize}
\setlength\itemsep{-0.1cm}
\item Repeatedly searches for causes and treatments related to specific body areas such as herniated discs in the lower back and neck, as well as pelvic pain.
\item Focuses on information about treatment options and costs, including medical braces, injection treatments.
\item Explores detailed management strategies including treatment reviews, practical tips, and rehabilitation methods for maintaining health during long working hours.
\end{itemize}
\vspace{-0.1cm}
\textbf{Evidence memory IDs:} 52, 61, 62, 63, 64, 65, 57, 67
\vspace{-0.2cm}
\begin{itemize}
\setlength\itemsep{-0.1cm}
\item \textbf{52.} Cause and solutions for lateral pelvic pain:  
The user repeatedly searches for the causes and solutions of pelvic-side pain, indicating an interest in understanding and resolving this physical issue.
\item \textbf{61.} Injection treatment for lumbar disc herniation:  
Searches and document views about injection therapy for lumbar disc herniation suggest that the user is exploring treatment options and making treatment decisions.
\item \textbf{62.} CT scan cost for lower back pain at a local orthopedic clinic: Searches related to CT scan costs for lower back pain at local orthopedic clinics indicate interest in medical examination costs.
\item \textbf{63.} Necessity and cost of lumbar CT scans:  
The user searches about whether CT scans are necessary and their associated costs, suggesting an effort to evaluate medical examination options.
\item \textbf{64.} Back health management during long working hours:  
Searches related to stretching for lower-body pain relief, back health maintenance, and rehabilitation during long working hours indicate interest in back health management.
\item \textbf{65.} Nerve injection treatment for cervical disc herniation:  
Searches about nerve injection treatments for cervical disc issues indicate interest in treatment methods for neck-related spinal problems.
\item \textbf{57.} Medical brand medical back brace (features, effectiveness, reviews):  
The user searches for product functionality and effectiveness of the medical brand back brace and refers to strong user recommendations.
\item \textbf{67.} Back brace usage methods and effectiveness:  
Repeated searches about how to wear a back brace, reasons for wearing it, and its effectiveness indicate interest in back health management and practical usage methods.
\end{itemize}
\end{tcolorbox}

\begin{tcolorbox}[colback=white,colframe=black,title=\texttt{\small{Low truthfulness (score = 0.2)}}]
\small
\textbf{Persona:} 
\vspace{-0.2cm}
\begin{itemize}
\setlength\itemsep{-0.1cm}
\item Self-improvement–oriented explorer seeking to improve daily routines and productivity.
\end{itemize}
\vspace{-0.1cm}
\textbf{Descriptions}
\vspace{-0.2cm}
\begin{itemize}
\setlength\itemsep{-0.1cm}
\item Regularly searches for self-improvement information at consistent times of the day.
\item Prioritizes study groups, structured learning systems, and performance-driven resources.
\item Prefers concrete and actionable methods to avoid inefficient use of time.
\end{itemize}
\vspace{-0.1cm}
\textbf{Evidence memory IDs:} 14, 15, 16
\vspace{-0.2cm}
\begin{itemize}
\setlength\itemsep{-0.1cm}
\item \textbf{14.} Online English learning system (free benefits):  
Search logs about free benefits for students enrolled in the online program indicate interest in an English learning system and study groups.
\item \textbf{15.} University admission information (2026 applications, regular admission tracks): Search logs related to applications for the 2026 admission cycle and regular admission tracks across several universities indicate interest in university admission information.
\item \textbf{16.} University rankings and global evaluation comparisons (2026 QS rankings, regional universities): Search logs about the 2026 QS World University Rankings and rankings of regional universities indicate interest in university evaluations and global ranking comparisons.
\end{itemize}
\end{tcolorbox}
\vspace{0.2cm}

\subsubsection{Qualitative Examples of Cluster Cohesion}
\label{appendix:cohesion_score}
Embedding-based cluster cohesion score measures how consistently the evidence memories within a cluster share a common behavioral theme. In the high-scoring example below, the memories revolve around a consistent topic related to childcare leave policies and financial benefits, forming a coherent behavioral theme. The induced persona summarizes this shared topic in a concise and well-supported way. In contrast, the low-scoring example mixes highly unrelated topics, and the induced persona becomes vague and less coherent.
\vspace{0.1cm}

\begin{tcolorbox}[colback=white,colframe=black,title=\texttt{\small{High cluster cohesion (score = 0.95)}}]
\small

\textbf{Persona:}
\vspace{-0.2cm}
\begin{itemize}
\setlength\itemsep{-0.1cm}
\item Practical information seeker with strong interest in childcare and household financial planning.
\end{itemize}

\vspace{-0.1cm}
\textbf{Descriptions}
\vspace{-0.2cm}
\begin{itemize}
\setlength\itemsep{-0.1cm}
\item Searches for detailed financial information related to parental leave, including compensation, take-home pay, tax deductions, and application procedures.
\item Shows interest in social support systems such as paternity leave and government childcare incentives.
\item Seeks a practical understanding of the legal and financial aspects of childcare-related support policies.
\end{itemize}

\vspace{-0.1cm}
\textbf{Evidence memory IDs:} 50, 53, 110, 132, 198, 199
\vspace{-0.2cm}
\begin{itemize}
\setlength\itemsep{-0.1cm}
\item \textbf{50.} Maternity leave and parental leave benefit information: Search logs about maternity leave and parental leave compensation indicate interest in childcare-related financial support.
\item \textbf{53.} How to apply for paternity leave benefits: Search activity about paternity leave and the application process for related benefits suggests interest in male participation in childcare and associated financial support.
\item \textbf{110.} Year-end tax deduction conditions for parental leave: The user searches for information about tax deduction conditions during parental leave, indicating interest in childcare-related financial policies and tax support.
\item \textbf{132.} Experiences of male parental leave usage: Search activity about men’s experiences using parental leave indicates interest in paternal childcare participation and related support policies.
\item \textbf{198.} Parental leave compensation and government incentive policies: Searches about parental leave compensation and government incentive policies suggest interest in financial support available during childcare.
\item \textbf{199.} Parental leave compensation and government incentive policies: Repeated searches about parental leave benefits and incentive programs indicate continued interest in childcare-related financial assistance.
\end{itemize}
\end{tcolorbox}

\begin{tcolorbox}[colback=white,colframe=black,title=\texttt{\small{Low cluster cohesion (score = 0.18)}}]
\small

\textbf{Persona:}
\vspace{-0.2cm}
\begin{itemize}
\setlength\itemsep{-0.1cm}
\item Exploratory information seeker who resolves a wide range of everyday questions through online search.
\end{itemize}

\vspace{-0.1cm}
\textbf{Descriptions}
\vspace{-0.2cm}
\begin{itemize}
\setlength\itemsep{-0.1cm}
\item Searches for practical but diverse information, including word meanings, sports news, PPT templates, transportation schedules, and appliance usage.
\item Tends to seek accurate and complete information to clarify small confusions or solve everyday problems.
\item Shows a broad preference for concrete answers and practical guidance across unrelated domains.
\end{itemize}

\vspace{-0.1cm}
\textbf{Evidence memory IDs:} 119, 126, 127, 128, 135, 142
\vspace{-0.2cm}
\begin{itemize}
\setlength\itemsep{-0.1cm}
\item \textbf{119.} Meaning and usage of the Korean words: Search activity aimed at accurately understanding and clarifying the meaning and usage of native Korean expressions.
\item \textbf{126.} Japan national football team and World Cup winning prospects: Search logs about the Japan national football team’s chances of winning the World Cup, including coach statements and AI-based predictions.
\item \textbf{127.} Korean national football team and coach-related issues: Search activity about the Korean national football team, player recruitment, basecamp issues, and coach-related controversies.
\item \textbf{128.} Express bus schedule planning for Geoje Gohyeon area: Search activity about bus schedules and fare tables for routes departing from Geoje Gohyeon terminal.
\item \textbf{135.} Information search about Trump’s negotiation strategies: Search logs indicating interest in negotiation techniques associated with Donald Trump.
\item \textbf{142.} Winter use of the WashTower dryer: Repeated searches about appliance use and maintenance, especially the use of a dryer during winter.
\end{itemize}
\end{tcolorbox}

\vspace{0.1cm}
\subsubsection{Effect of Training Data Size}
\label{appendix:data_size_exp}

\begin{figure*}[t!]
    \centering
    \includegraphics[width=1.0\linewidth]{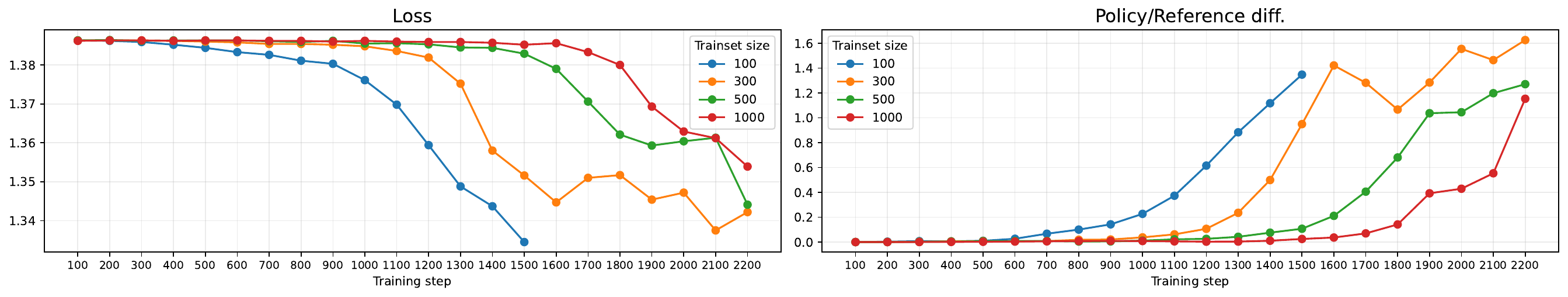}
    \caption{Training dynamics under different training data sizes on the \texttt{Srv.} dataset. The plots show the training loss and the policy--reference log-probability difference during optimization.}
    \label{fig:train_log}
\end{figure*}

\begin{figure*}[h!]
    \centering
    \includegraphics[width=1.0\linewidth]{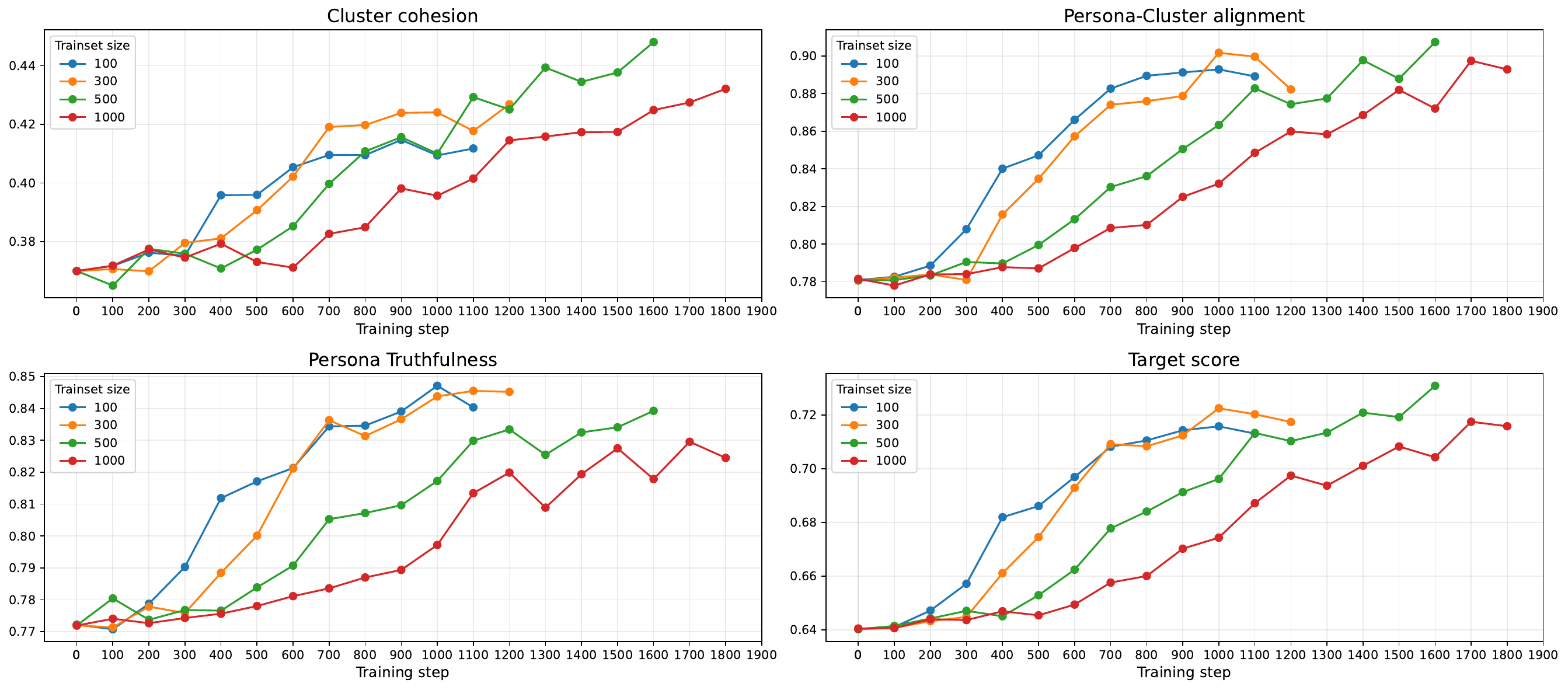}
    \caption{Validation metrics under different training data sizes on the \texttt{Srv.} dataset. We report cluster cohesion, persona--cluster alignment, persona truthfulness, and the final target score during training. Increasing the training size improves persona quality up to around 500 users, after which the gains become limited. Each configuration is shown only up to the point where unparseable validation outputs exceed 5\%.}
    \label{fig:val_log}
\end{figure*}

We analyze the effect of training data size on the \texttt{Srv.} dataset by varying the number of training users and monitoring both training dynamics and validation performance. Figure~\ref{fig:train_log} shows the training logs, including loss and policy--reference differences, while Figure~\ref{fig:val_log} reports validation metrics related to persona quality. Across different training sizes, optimization progresses as expected, and the validation metrics generally improve during training before plateauing or fluctuating at later stages. Increasing the training size from 100 to 300 and 500 users leads to noticeable improvements in validation metrics such as cluster cohesion, persona--cluster alignment, and truthfulness. However, the gains become marginal beyond 500 users. In particular, the target score and individual quality metrics show only marginal improvements when increasing the training size from 500 to 1000 users.

In longer training runs, some models began to produce malformed outputs on the validation set, such as format violations or unparseable generations. Because a small portion of such failures can also arise from overly long outputs being truncated, we use a relatively tolerant stopping criterion and stop training once the proportion of unparseable validation outputs exceeds 5\%. Based on these observations, our experimental setup does not appear to require a large amount of training data to achieve strong performance. We therefore fix the training size to 500 users for all datasets in the main experiments, as it provides a good trade-off between computational efficiency and persona quality.

\subsubsection{Induced Personas}
\label{appendix:example-lt}
This section provides additional examples of personas induced by the initial policy $\pi_0$ and the trained policy $\pi_\theta$ across datasets. The examples illustrate how training can improve the specificity and evidence grounding of the generated personas. Figures~\ref{fig:persona_example_merrec} and \ref{fig:persona_example_aol} show representative examples from the \texttt{MerRec} and \texttt{AOL} datasets, respectively. 

\begin{figure}[htp!]
\centering
\begin{tcolorbox}[colback=white,colframe=black,title=\texttt{\small{Initial policy ($\pi_0$)}}]
\small
\vspace{-0.1cm}
\textbf{Persona:}
\vspace{-0.2cm}
\begin{itemize}
\setlength\itemsep{-0.1cm}
\item Dedicated Auto Enthusiast
\end{itemize}

\vspace{-0.2cm}
\textbf{Descriptions}
\vspace{-0.2cm}
\begin{itemize}
\setlength\itemsep{-0.1cm}
\item Focus on maintaining and upgrading a specific vehicle (Honda Civic).
\item Explores both essential replacement parts and performance enhancements.
\item Demonstrates interest in vintage and modern components.
\end{itemize}

\vspace{-0.2cm}
\textbf{Evidence memory IDs:} 11, 12, 14, 15, 54, 60
\vspace{-0.1cm}
\end{tcolorbox}

\begin{tcolorbox}[colback=white,colframe=black,title=\texttt{\small{Trained policy ($\pi_\theta$)}}]
\small
\vspace{-0.1cm}
\textbf{Persona:}
\vspace{-0.2cm}
\begin{itemize}
\setlength\itemsep{-0.1cm}
\item Honda Civic enthusiast.
\end{itemize}

\vspace{-0.2cm}
\textbf{Descriptions}
\vspace{-0.2cm}
\begin{itemize}
\setlength\itemsep{-0.1cm}
\item Focus on Honda Civic parts replacement and vintage accessories.
\item Shows particular interest in early-2000s Civic models and related components such as LED headlight assemblies.
\item Explores maintenance, repair, and upgrade options including audio and performance parts.
\end{itemize}

\vspace{-0.2cm}
\textbf{Evidence memory IDs:} 11, 12, 14, 15, 26, 54, 60, 61, 64
\vspace{-0.1cm}
\end{tcolorbox}
\caption{Generated personas before and after training with largely overlapping evidence memories, illustrated with an example from the \texttt{MerRec} dataset.}
\label{fig:persona_example_merrec}
\end{figure}

\begin{figure}[htp!]
\centering
\begin{tcolorbox}[colback=white,colframe=black,title=\texttt{\small{Initial policy ($\pi_0$)}}]
\small
\vspace{-0.1cm}
\textbf{Persona:}
\vspace{-0.2cm}
\begin{itemize}
\setlength\itemsep{-0.1cm}
\item Value-Conscious Homebody
\end{itemize}

\vspace{-0.2cm}
\textbf{Descriptions}
\vspace{-0.2cm}
\begin{itemize}
\setlength\itemsep{-0.1cm}
\item Prioritizes frugal living and staying at home.
\item Actively seeks discounts and deals.
\item Explores home goods and personal care products through online retailers.
\end{itemize}

\vspace{-0.2cm}
\textbf{Evidence memory IDs:} 11, 12, 13, 14, 29, 30, 45, 46, 62
\vspace{-0.1cm}
\end{tcolorbox}

\begin{tcolorbox}[colback=white,colframe=black,title=\texttt{\small{Trained policy ($\pi_\theta$)}}]
\small
\vspace{-0.1cm}
\textbf{Persona:}
\vspace{-0.2cm}
\begin{itemize}
\setlength\itemsep{-0.1cm}
\item Frugal \& Practical Home Manager
\end{itemize}

\vspace{-0.2cm}
\textbf{Descriptions}
\vspace{-0.2cm}
\begin{itemize}
\setlength\itemsep{-0.1cm}
\item Interest in frugal living and staying at home (cost reduction).
\item Exploration of tax exemptions and financial services.
\item Shopping for home goods and utilizing coupons (Home Depot, QVC, HSN).
\end{itemize}

\vspace{-0.2cm}
\textbf{Evidence memory IDs:} 11, 12, 13, 14, 16, 29, 30, 38, 39, 49
\vspace{-0.1cm}
\end{tcolorbox}
\caption{Generated personas before and after training with largely overlapping evidence memories, illustrated with an example from the \texttt{AOL} dataset.}
\label{fig:persona_example_aol}
\end{figure}

\subsection{Experimental Details}
\label{appendix:exp_details}

\subsubsection{Ours ($\pi_\theta$)}
\label{appendix:exp_detail_ours}

\paragraph{Policy parameterization.}
For each backbone model, we initialize the trainable persona policy $\pi_\theta$ from the corresponding base model $\pi_0$. A frozen copy of the same model is used as the reference policy $\pi_{\mathrm{ref}}$ in the groupwise DPO objective. We fine-tune the policy using Low-Rank Adaptation (LoRA) \cite{hu2022lora}, where only the adapter parameters are updated while the backbone model weights remain frozen. The LoRA adapters use rank $r=16$, scaling factor $\alpha=32$, and dropout rate $0.05$. Training is performed in float16 precision without quantization.

\paragraph{Training instances and prompt format.}
Each training instance corresponds to a user window $\mathbf{W}_t$, represented as a chronologically ordered list of intent memories. The input prompt contains task instructions and the list of intent memories extracted from daily logs. The model is prompted to produce a structured set of personas, where each persona consists of a persona label, $K$ supporting descriptions, and an evidence set referencing the assigned memories. The full prompt template is provided in Appendix~\ref{appendix:prompt-lt}.

\paragraph{Offline candidate generation.}
For each training window $\mathbf{W}_t$, we generate $n=8$ candidate outputs from the initial policy $\pi_0$ offline prior to training and reuse them throughout optimization. To encourage output diversity, we use sampling-based decoding with temperature $0.9$ and top-p $=0.9$. The total context length (input + output) is capped at 30,720 tokens. Each generated output is then parsed into a structured persona set. Outputs that do not follow the required json format or cannot be successfully parsed are discarded before training. We found that retaining malformed outputs with default low rewards was undesirable in our groupwise objective, because such outputs still receive non-zero probability mass within the softmax normalization and can therefore contribute to the learning signal. By removing unparsable candidates entirely, we ensure that training is driven only by valid structured persona outputs.

\begin{table*}[t!]
\centering
\small
\resizebox{1.0\linewidth}{!}{%
\begin{tabular}{ccccccccc}
\toprule
\multicolumn{4}{c}{\textbf{\texttt{Srv.}}} &
\multicolumn{4}{c}{\textbf{\texttt{MerRec}}} &
\multicolumn{1}{c}{\textbf{\texttt{AOL}}} \\
\cmidrule(lr){1-4} \cmidrule(lr){5-8} \cmidrule(lr){9-9}
Qwen3-14B & Qwen3-30B & Gemma3-12B & Gemma3-27B & Qwen3-14B & Qwen3-30B & Gemma3-12B & Gemma3-27B & Gemma3-27B \\
\midrule
1,100 & 1,600 & 600 & 900 & 900 & 1,600 & 700 & 800 & 1,100 \\
\bottomrule
\end{tabular}%
}
\caption{Best checkpoint step selected on the validation set for each backbone model and dataset.}
\vspace{-0.3cm}
\label{tab:best_checkpoint_steps}
\end{table*}

\paragraph{Reward computation.}
Persona-level rewards are computed as described in Section~\ref{sec:method}. The cohesion score combines an intra-cluster cohesion term and a separation term from non-evidence memories, with a intra-variance penalty controlled by $\lambda=0.5$. Evidence-set sizes are constrained using parameters $e_{\min}=3$ and $e_{\max}=20$. When a persona contains fewer than two evidence memories, the intra-cluster component is omitted and the size constraint discourages degenerate clusters. Alignment and truthfulness are computed using an LLM-based judge. The judge is prompted to output structured scores in the range $[0,1]$ for each evaluation. For truthfulness, we score both the persona label and each supporting description and average the results.

\paragraph{Groupwise DPO optimization.}
We pre-compute scalar rewards for all candidate outputs. We set the reward weights to $\alpha_1=0.9$, $\alpha_2=0.1$, $\alpha_3=0.9$, and $\alpha_4=0.1$, placing emphasis on persona-level quality signals while treating cluster size as a regularizer and coverage as a global constraint. During training, we repeatedly resample a subset of size $G=4$ from the candidate pool for each window $\mathbf{W}_t$ across updates and construct a soft preference distribution over the selected candidates. The policy $\pi_\theta$ is optimized to match this preference distribution using the groupwise DPO objective described in Section~\ref{subsec:method-optimization}. The sharpness of preference amplification is controlled by the parameter $\beta=0.5$, and a deviation penalty weighted by $\lambda_{\mathrm{KL}}=0.005$ stabilizes training. Sequence log-probabilities are computed as the mean of token-level log-probabilities over the generated output, excluding padded tokens.

\paragraph{Optimization details.}
We train the model using the AdamW optimizer \cite{loshchilov2018decoupled} with a learning rate of $1\times10^{-5}$, $\beta_1=0.9$, $\beta_2=0.95$, $\epsilon=10^{-8}$, and no weight decay. Training is performed with batch size $1$. For each training instance, a candidate group of size $G=4$ is sampled for the groupwise DPO objective. We use a warmup phase for the first $5\%$ of training steps, followed by a cosine learning-rate schedule. Training runs for 30 epochs, and the total number of optimizer steps is determined by the number of training windows. We evaluate the model on the validation set every 100 steps and select the final checkpoint based on the highest persona quality score on the validation set. The selected checkpoint step for each backbone model and dataset is reported in Table~\ref{tab:best_checkpoint_steps}.

\paragraph{Hardware and Inference}
All experiments are conducted on NVIDIA H100 GPUs (80GB). Each backbone model is trained independently using two GPUs with LoRA fine-tuning. For validation and test evaluation, we use the vLLM \cite{kwon2023efficient} inference engine to load the backbone models together with the corresponding trained LoRA adapters for efficient generation.

\subsubsection{PersonaX}
\label{appendix:exp_detail_personaX}

We implement PersonaX-style baselines following the core design of \citet{shi2025personax}, adapted to our inputs and evaluation protocol. Since our framework operates on intent-level memories, we apply PersonaX's clustering and persona labeling pipeline to intent memories instead of items. For all LLM calls in both variants, we use Gemma3-27B. Below, we describe the adaptation details for each component.

\paragraph{Embedding model.}
We utilize the same embedding models as our method: BGE-M3-ko for the Korean \texttt{Srv.} dataset and BGE-M3 for the English \texttt{MerRec} and \texttt{AOL} datasets. Each intent memory is embedded by concatenating its memory label and description fields into a single string of the form \texttt{"\{intent\_memory\}: \{description\}"}, and the resulting embeddings are L2-normalized.

\paragraph{Clustering hyperparameters.}
We follow PersonaX's hierarchical agglomerative clustering with Ward linkage and Euclidean distance. We set the distance threshold to $\tau=1.4$ after a grid search over $[0.5, 1.5]$; lower values produced predominantly singleton clusters under the BGE-M3 embedding space, rendering the clustering ineffective. For the prototypicality--diversity trade-off parameter, we use $\alpha=1.07$, the midpoint of the recommended range $[1.06, 1.08]$ from the original paper. The sampling budget ratio, which controls the overall fraction of memories selected as representative samples, is set to $0.7$ to match the coverage soft threshold used in our reward formulation (Section~\ref{sec:method}).

\paragraph{$\text{PersonaX}_s$ (Summarization).}
We use our persona induction prompt (Appendix~\ref{appendix:prompt-lt}), modifying the instruction to generate exactly one persona per cluster to match their methodology.


\paragraph{$\text{PersonaX}_r$ (Reflection).}
Following PersonaX, we apply the AgentCF-style sequential reflection pipeline. Starting from an empty profile (``Currently Unknown''), the model iteratively refines its persona by choosing between positive and negative memory candidates for the user and reflecting on incorrect choices.

\paragraph{Negative sampling.}
For each positive memory within a cluster, we draw a negative memory from other clusters of intermediate similarity to the source cluster, yielding more informative negatives. Specifically, we represent each cluster by its centroid, computed as the L2-normalized mean of the selected intent-memory embeddings. We then randomly select up to 10 clusters from other users whose centroid cosine similarity to the source cluster falls within $[0.5, 0.85]$. If fewer than 10 clusters fall within this range, we fall back to the top-10 most similar clusters from other users. Memories from the selected clusters serve as the negative pool, from which we randomly sample one negative memory per positive memory.

\paragraph{Inference and reflection.}
At each step, the model receives its current persona profile and a pair of memories (one positive, one negative) and must choose the preferred one with an explanation. If the model chooses incorrectly, it is prompted to reflect on the mistake and update its profile. We set the maximum number of retries per pair to 1 and enforce a reflection on the first pair to bootstrap the persona from the empty initial state. The full prompts are provided in Figures~\ref{fig:prompt_personax_a_infer} and~\ref{fig:prompt_personax_a_reflect}.

\subsection{Examples}
\label{appendix:example}

\subsubsection{Daily Logs from Each Dataset}
\label{appendix:example-log}
Figures~\ref{fig:example_log_merrec} and~\ref{fig:example_log_aol} show daily logs from \texttt{MerRec} and \texttt{AOL}, respectively. For the proprietary \texttt{Srv.} dataset, raw user logs cannot be publicly disclosed due to privacy and confidentiality constraints. Therefore, instead of showing the original logs, we provide an excerpt of the corresponding intent-level memories derived from them in Section \ref{appendix:example-st}.

\subsubsection{From Daily Logs to Intent-level Memories}
\label{appendix:example-st}
Figures~\ref{fig:example_st_merrec} and~\ref{fig:example_st_aol} show intent-level memories derived from the example daily logs presented in Section \ref{appendix:example-log}. Figure~\ref{fig:example_st_srv} presents an excerpt of intent-level memories from the \texttt{Srv.} dataset, for which the corresponding raw user logs are not shown.

\subsection{Prompts}
\label{appendix:prompt}

\subsubsection{Intent-Level Memory Summarization Prompt}
\label{appendix:prompt-st}
Figure~\ref{fig:prompt_st} shows the prompt used to summarize daily user logs into intent-level memories. Given raw behavioral logs, the prompt instructs the LLM to identify exploration intents and generate structured intent memories consisting of a concise intent phrase and a grounded description.

\subsubsection{Persona Induction Prompt}
\label{appendix:prompt-lt}
Figure~\ref{fig:prompt_lt} shows the prompt used to induce personas from intent-level memories. The prompt instructs to identify recurring behavioral patterns and lifestyle tendencies, and to organize them into personas supported by evidence memories.

\subsubsection{Judge Prompts}
\label{appendix:prompt-judge}
Figures~\ref{fig:prompt_judge_alignment} and~\ref{fig:prompt_judge_truth} show the prompts used for LLM-based evaluation of persona quality, corresponding to the alignment and truthfulness, respectively.

\clearpage
\onecolumn

\hspace{-0.5cm} \textbf{Examples}

\begin{tcolorbox}[breakable=false, width=\textwidth, colback=white, colframe=black, title=\texttt{\small{\# MerRec. User:27011435 Date:2023-05-12}}]
\small
    
    User viewed [Men, Tops, T-shirts] Cactus Jack Fragment shirt
    
    User liked [Men, Tops, T-shirts] Cactus Jack Fragment shirt
    
    User viewed [Women, Tops \& blouses, T-shirts] KAWS x Uniqlo Companion T-Shirt
    
    User viewed [Men, Tops, T-shirts] travis scott kaws shirt
    
    User liked [Men, Tops, T-shirts] travis scott kaws shirt
    
    User viewed [Men, Men's accessories, Hats] Hat club exclusive OG poly China Sandstorm new era fitted 7 1/4
    
    User viewed [Men, Sweats \& hoodies, Sweat Pants] joggers lululemon  mens
    
    User viewed [Men, Athletic apparel, Shorts] Lululemon Fast and Free 
    Lined Short 6” Men’s Medium Blue Chill NWT
    
    User viewed [Men, Tops, T-shirts] 2020 cactus jack t-shirt
    
    User viewed [Men, Tops, T-shirts] Supreme t shirt
    
    User viewed [Toys \& Collectibles, Sports Trading Cards] Captain Mariano Rivera Bobblehead
    
    User viewed [Toys \& Collectibles, Collectibles \& Hobbies, Souvenirs \& Memorabilia] New York Yankees Bobblehead
    
    User viewed [Toys \& Collectibles, Action Figures \& Accessories, Action Figures] Breaking Bad Collectible Figure
    
    User viewed [Men, Shoes, Athletic] Nike Air Skylon 2 Fear of God
    
    User viewed [Men, Sweats \& hoodies, Sweat Pants] Shadow hill sweatpants
    
    User viewed [Women, Shoes, Athletic] New Balance 990v5
    
    User viewed [Kids, Boys shoes, Boys 2T-5T] New balance 990
    
    User viewed [Men, Shoes, Athletic] New balance 990 v4  mens sz 11
    
    User viewed [Men, Shoes, Athletic] air force 1 white
    
    User liked [Men, Shoes, Athletic] Nike Dunk high
    
    User liked [Men, Shoes, Athletic] air force 1 white
    
    User viewed [Electronics, Headphones \& MP3 Players, Bluetooth Headphones] Apple AirPods Pro 1st Gen
    
    User viewed [Women, Jeans, Skinny Jeans] Ksubi jeans
    
    User viewed [Men, Shoes, Fashion sneakers] Yeezy 350s
    
    User viewed [Men, Shoes, Athletic] Air Force 1
    
    User viewed [Men, Tops, T-shirts] travis scott kaws shirt
    
    User viewed [Vintage \& collectibles, Antique, Collectibles] Kaws Reese’s Puffs
    
    User liked [Vintage \& collectibles, Antique, Collectibles] Kaws Reese’s Puffs
    
    User viewed [Beauty, Makeup, Nails] Yellow kaws press ons
    
    User viewed [Men, Shoes, Athletic] air force 1 white
    
    \end{tcolorbox}
\vspace{-0.2cm}
\captionof{figure}{Example daily activity log from the \texttt{MerRec} dataset, showing a sequence of user interactions with items, including views and likes across product categories.}
\vspace{0.5cm}
\label{fig:example_log_merrec}

\begin{tcolorbox}[width=\textwidth, colback=white, colframe=black, title=\texttt{\small{\# AOL. User:6680310 Date:2006-03-20}}]
\small    
    User searched italian greyhound videos

    User searched pomeranian video

    User searched french bulldog videos

    User searched cabrillo college

    User searched copper canisters

    User searched kitchen canisters

    User searched grandin road

    User searched sur la table

    User searched utensil holder

    User searched saint mary's college of california

    User searched cabrillo college

    User searched french bulldog videos
    
    \end{tcolorbox}
\vspace{-0.2cm}
\captionof{figure}{Example daily search log from the \texttt{AOL} dataset, illustrating a sequence of user search queries within a single day.}
\vspace{0.5cm}
\label{fig:example_log_aol}

\begin{tcolorbox}[width=\textwidth, colback=white, colframe=black, title=\texttt{\small{\# MerRec. UserID:27011435 Date:2023-05-12}}, breakable]
\small

[ 

\hspace{0.1em} \{ 

\hspace{0.3em} "intent\_memory": "Cactus Jack Fragment shirts (Men's T-shirts + Purchase Consideration)", 

\hspace{0.4em} "description": "The user repeatedly viewed and liked Cactus Jack Fragment shirts, indicating a strong interest in 

\hspace{6.5em} purchasing or learning more about this specific item. The repeated views and likes suggest 

\hspace{6.5em} a focused exploration intent." 

\hspace{0.1em} \}, 

\hspace{0.1em} \{

\hspace{0.3em} "intent\_memory": "Travis Scott KAWS collaborations (Men's T-shirts + Collection/Purchase)", 

\hspace{0.4em} "description": "The user viewed and liked multiple Travis Scott KAWS shirts, demonstrating an interest in this specific 

\hspace{6.5em} collaboration line. The repeated views and likes suggest a focused exploration intent." 

\hspace{0.1em} \}, 

\hspace{0.1em} \{ 

\hspace{0.3em} "intent\_memory": "New Era fitted hats (Men's Accessories + Style Exploration)", 

\hspace{0.4em} "description": "The user viewed a specific New Era fitted hat, suggesting an interest in exploring men's accessories 

\hspace{6.5em} and potentially specific hat styles." 

\hspace{0.1em} \}, 

\hspace{0.1em} \{ 

\hspace{0.3em} "intent\_memory": "Lululemon athletic wear (Men's athletic apparel + Purchase Consideration)", 

\hspace{0.4em} "description": "The user viewed joggers and shorts from Lululemon, indicating an interest in athletic apparel from 

\hspace{6.5em} this brand. The views suggest a consideration for purchase." 

\hspace{0.1em} \}, 

\hspace{0.1em} \{ 

\hspace{0.3em} "intent\_memory": "Bobbleheads (Sports Collectibles + Collecting)",

\hspace{0.4em} "description": "The user viewed both a Captain Mariano Rivera and a New York Yankees bobblehead, 

\hspace{6.5em} indicating an interest in sports memorabilia and collecting." 

\hspace{0.1em} \}, 

\hspace{0.1em} \{ 

\hspace{0.3em} "intent\_memory": "Breaking Bad collectible figures (Action Figures + Collecting)", 

\hspace{0.4em} "description": "The user viewed a Breaking Bad collectible figure, suggesting an interest in action figures 

\hspace{6.5em} and potentially collecting items related to the show." 

\hspace{0.1em} \}, 

\hspace{0.1em} \{ 

\hspace{0.3em} "intent\_memory": "Nike Air Skylon 2 Fear of God (Men's Athletic Shoes + Style Exploration)", 

\hspace{0.4em} "description": "The user viewed this specific Nike shoe, indicating an interest in athletic shoe styles 

\hspace{6.5em} and potentially collaborations." 

\hspace{0.1em} \}, 

\hspace{0.1em} \{ 

\hspace{0.3em} "intent\_memory": "Shadow Hill sweatpants (Men's Sweat Pants + Style Exploration)", 

\hspace{0.4em} "description": "The user viewed Shadow Hill sweatpants, suggesting an interest in sweatpants and potentially specific 

\hspace{6.5em} brands or styles." 

\hspace{0.1em} \}, 

\hspace{0.1em} \{ 

\hspace{0.3em} "intent\_memory": "New Balance 990 series (Men's and Kids' Athletic Shoes + Purchase Consideration)", 

\hspace{0.4em} "description": "The user viewed multiple New Balance 990 variations (v5 for women, and v4 for men), including 

\hspace{6.5em} a kids' size, indicating a broad interest in this shoe series and potentially purchasing for themselves 

\hspace{6.5em} and their children." 

\hspace{0.1em} \}, 

\hspace{0.1em} \{ 

\hspace{0.3em} "intent\_memory": "Air Force 1 (Men's Athletic Shoes + Style Exploration)", 

\hspace{0.4em} "description": "The user repeatedly viewed and liked Air Force 1 shoes, indicating a strong interest 

\hspace{6.5em} in this classic sneaker style." 

\hspace{0.1em} \}, 

\hspace{0.1em} \{ 

\hspace{0.3em} "intent\_memory": "Apple AirPods Pro 1st Gen (Bluetooth Headphones + Purchase Consideration)", 

\hspace{0.4em} "description": "The user viewed Apple AirPods Pro, suggesting an interest in Bluetooth headphones 

\hspace{6.5em} and potentially considering a purchase." 

\hspace{0.1em} \}, 

\hspace{0.1em} \{ 

\hspace{0.3em} "intent\_memory": "Ksubi jeans (Women's Skinny Jeans + Style Exploration)", 

\hspace{0.4em} "description": "The user viewed Ksubi jeans, indicating an interest in women's denim and potentially specific brands 

\hspace{6.5em} or styles." 

\hspace{0.1em} \}, 

\hspace{0.1em} \{ 

\hspace{0.3em} "intent\_memory": "Yeezy 350s (Men's Fashion Sneakers + Style Exploration)", 

\hspace{0.4em} "description": "The user viewed Yeezy 350s, indicating an interest in fashion sneakers and potentially specific brands 

\hspace{6.5em} or styles." 

\hspace{0.1em} \}, 

\hspace{0.1em} \{ 

\hspace{0.3em} "intent\_memory": "Kaws Reese’s Puffs (Vintage Collectibles + Collecting)", 

\hspace{0.4em} "description": "The user viewed and liked Kaws Reese’s Puffs, indicating an interest in vintage collectibles 

\hspace{6.5em} and potentially specific collaborations." 

\hspace{0.1em} \}, 

\hspace{0.1em} \{ 

\hspace{0.3em} "intent\_memory": "Yellow Kaws press ons (Beauty, Nails + Style Exploration)", 

\hspace{0.4em} "description": "The user viewed yellow Kaws press ons, indicating an interest in beauty products and potentially 

\hspace{6.5em} specific collaborations." 

\hspace{0.1em} \}, 

\hspace{0.1em} \{ 

\hspace{0.3em} "intent\_memory": "Supreme t shirts (Men's T-shirts + Style Exploration)", 

\hspace{0.4em} "description": "The user viewed Supreme t shirts, indicating an interest in men's t-shirts and potentially specific brands 

\hspace{6.5em} or styles." 

\hspace{0.1em} \} 

]
    
\end{tcolorbox}
\vspace{-0.2cm}
\captionof{figure}{Example intent-level memories generated from the daily activity log in Figure~\ref{fig:example_log_merrec} (\texttt{MerRec} dataset).}
\vspace{0.5cm}
\label{fig:example_st_merrec}

\begin{tcolorbox}[breakable, width=\textwidth, colback=white, colframe=black, title=\texttt{\small{\# AOL. UserID:6680310 Date:2006-03-20}}]
\small
[ 

\hspace{0.1em} \{

\hspace{0.3em} "intent\_memory": "French Bulldog videos (entertainment and observation)", 

\hspace{0.4em} "description": "The user searched for 'french bulldog videos,' indicating a clear interest in observing and 

\hspace{6.5em} enjoying content related to this dog breed. The repetition reinforces this as a primary interest." 

\hspace{0.1em} \}, 

\hspace{0.1em} \{ 

\hspace{0.3em} "intent\_memory": "Cabrillo College (academic exploration)",

\hspace{0.4em} "description": "The user searched for 'Cabrillo College', suggesting an exploration of the college, 

\hspace{6.5em} potentially for academic purposes like admissions or course information." 

\hspace{0.1em} \}, 

\hspace{0.1em} \{

\hspace{0.3em} "intent\_memory": "Kitchen canisters (home decor and organization)", 

\hspace{0.4em} "description": "The user searched for 'kitchen canisters' and 'copper canisters,' indicating an interest in kitchen 

\hspace{6.5em} storage and decorative items. The inclusion of 'copper canisters' suggests a preference for 

\hspace{6.5em} a specific material." 

\hspace{0.1em} \}, 

\hspace{0.1em} \{ 

\hspace{0.3em} "intent\_memory": "Utensil holder (kitchen organization)", 

\hspace{0.4em} "description": "The user searched for 'utensil holder', indicating an interest in kitchen organization 

\hspace{6.5em} and storage solutions." 

\hspace{0.1em} \}, 

\hspace{0.1em} \{ 

\hspace{0.3em} "intent\_memory": "Saint Mary's College of California (academic exploration)", 

\hspace{0.4em} "description": "The user searched for 'Saint Mary's College of California,' suggesting an exploration of the college, 

\hspace{6.5em} potentially for academic purposes like admissions or course information." 

\hspace{0.1em} \}, 

\hspace{0.1em} \{ 

\hspace{0.3em} "intent\_memory": "Grandin Road (home goods shopping)",

\hspace{0.4em} "description": "The user searched for 'Grandin Road,' indicating an interest in browsing or purchasing home goods 

\hspace{6.5em} from this retailer." 

\hspace{0.1em} \}, 

\hspace{0.1em} \{ 

\hspace{0.3em} "intent\_memory": "Sur La Table (kitchenware shopping)", 

\hspace{0.4em} "description": "The user searched for 'Sur La Table,' indicating an interest in browsing or purchasing kitchenware 

\hspace{6.5em} from this retailer." 

\hspace{0.1em} \} 

]
    
\end{tcolorbox}
\vspace{-0.2cm}
\captionof{figure}{Example intent-level memories generated from the daily search log in Figure~\ref{fig:example_log_aol} (\texttt{AOL} dataset).}
\vspace{0.5cm}
\label{fig:example_st_aol}

\begin{tcolorbox}[width=\linewidth, colback=white, colframe=black, title=\texttt{\small{\# Srv. UserID:[Redacted] Date:2026.01.08}}, breakable]
\small

[ 

\hspace{0.1em} \{ 

\hspace{0.3em} "intent\_memory": "Miss Korea actress life stories in the U.S. (celebrity lifestyle exploration)", 

\hspace{0.4em} "description": "The user searched for stories about Miss Korea–origin actresses living in the United States, 

\hspace{6.5em} including topics such as parenting, dieting, and their past experiences. Queries such as 

\hspace{6.5em} 'Miss Korea actress living in a U.S. mansion' and 'Miss Korea actress U.S. mansion one-week diet' 

\hspace{6.5em} suggest an interest in celebrity lifestyles and personal narratives behind their public image." 

\hspace{0.1em} \}, 

\hspace{0.1em} \{ 

\hspace{0.3em} "intent\_memory": "Max Mara outlet shopping in Jiyugaoka, Tokyo (travel shopping planning)", 

\hspace{0.4em} "description": "The user searched for 'Max Mara outlet shopping mall in Jiyugaoka, Tokyo,' suggesting 

\hspace{6.5em} preparation for shopping during travel. This indicates an interest in locating specific 

\hspace{6.5em} brand outlets and planning a shopping itinerary." 

\hspace{0.1em} \}, 

\hspace{0.1em} \{ 

\hspace{0.3em} "intent\_memory": "Global popularity of Korean fruit soju (cultural trend observation)", 

\hspace{0.4em} "description": "The user searched for articles about Korean fruit-flavored soju becoming popular among 

\hspace{6.5em} young people in the United States, despite declining popularity in Korea. This suggests 

\hspace{6.5em} curiosity about international reactions to Korean products and cultural trends." 

\hspace{0.1em} \}, 

\vspace{2em}
\hspace{23em} . . .
\vspace{2em}

\hspace{0.1em} \{ 

\hspace{0.3em} "intent\_memory": "Oyster rice cake soup recipes (Korean cooking exploration)", 

\hspace{0.4em} "description": "The user searched for recipes such as 'how to cook oyster rice cake soup' and 

\hspace{6.5em} 'seaweed oyster rice cake soup recipe,' indicating an interest in preparing traditional 

\hspace{6.5em} Korean dishes, particularly variations of tteokguk." 

\hspace{0.1em} \}, 

\hspace{0.1em} \{ 

\hspace{0.3em} "intent\_memory": "Pistachio desserts and health benefits (food and nutrition interest)", 

\hspace{0.4em} "description": "The user searched for information about the health benefits of pistachios appearing in 

\hspace{6.5em} trendy green desserts, such as chewy cookies. This suggests an interest in both food trends 

\hspace{6.5em} and the nutritional value of ingredients." 

\hspace{0.1em} \}, 

\hspace{0.1em} \{ 

\hspace{0.3em} "intent\_memory": "Ethical issues in pet euthanasia (animal welfare concern)", 

\hspace{0.4em} "description": "The user searched for a news story about a dog that escaped a fire while the owner 

\hspace{6.5em} requested euthanasia. This indicates an interest in ethical debates and emotional 

\hspace{6.5em} stories related to animal welfare." 

\hspace{0.1em} \}, 

\hspace{0.1em} \{ 

\hspace{0.3em} "intent\_memory": "Foods that prevent blood sugar spikes (health management)", 

\hspace{0.4em} "description": "The user searched for 'Top 5 foods that prevent blood sugar spikes,' suggesting 

\hspace{6.5em} an interest in diet and nutrition related to diabetes prevention or blood sugar control." 

\hspace{0.1em} \}, 

\hspace{0.1em} \{ 

\hspace{0.3em} "intent\_memory": "Magnet behavior at corners (scientific curiosity)", 

\hspace{0.4em} "description": "The user searched for the phenomenon of attaching a magnet to a corner, suggesting 

\hspace{6.5em} curiosity about a physical or scientific principle related to magnetism." 

\hspace{0.1em} \} 

]
    
\end{tcolorbox}
\vspace{-0.2cm}
\captionof{figure}{Example intent-level memories from the proprietary \texttt{Srv.} dataset.}
\vspace{0.7cm}
\label{fig:example_st_srv}

\textbf{Prompts}

\begin{tcolorbox}[width=\textwidth, colback=white, colframe=black, title=\texttt{\small{\# Intent-Level Memory Summarization Prompt}}]
\small

    Based on the user logs provided, identify all topics of interest of the analyzed subject and organize them into an intent\_memory list. \\
    
    \textbf{Instructions} \\
    1. Understand the analyzed subject’s exploration flow and generate interest\_profile candidates:
    
    \hspace{1em} - Read all logs and collect all core topics that became exploration targets of the analyzed subject.
    
    \hspace{1em} - For each topic, identify from what perspective, context, and flow the subject is exploring it. 
    
    \hspace{1em} - Based on this, list up intent\_memory candidates.

    2. Consolidate and refine selected candidates:

    \hspace{1em} - If N intent\_memory candidates share the same intent or have clearly common characteristics, merge them into one.
    
    \hspace{1em} - If one intent\_memory contains N contents with different intents, split them into N separate memories.
    
    \hspace{1em} - Ensure that each intent\_memory is clearly distinguishable and has no overlap (no duplication). \\

    \textbf{Rules} \\
    An intent\_memory represents one exploration intent (One Intent) as a specific topic. Each item must be a noun phrase that captures detailed context.
    
    \hspace{1em} - Target: A clearly defined entity (proper noun / clearly identifiable concept).

    \hspace{1em} - Purpose/Angle: The reason or perspective from which the target is being explored.

    \hspace{1em} - Sub-dimensions (optional): Detailed contextual information. \\

    \textbf{Important notes:}

    \hspace{1em} - Be as specific as possible.

    \hspace{1em} - Include proper nouns whenever possible.

    \hspace{1em} - Only interpretations grounded in the logs are allowed (no excessive inference). \\

    \textbf{Output Format:} \\
    Return only a single, parsable JSON object in the following format:

    [
    
        \hspace{1em} \{
    
            \hspace{2em} "intent\_memory": "Interest topic title (Target + Purpose/Angle + Sub-dimensions, noun phrase)",
            
            \hspace{2em} "description": "A sentence to describe core keywords/topics and the inferred exploration context" \\

        \hspace{1em} \},
        
        \hspace{1em} ...

    ] \\

    Before output, verify that each intent\_memory includes a concrete target and context, remove duplicates, and then produce the final result. \\

    \textbf{User Logs:} \\
    \textcolor{gray}{\{\{ user logs \}\}}
    \end{tcolorbox}
\vspace{-0.2cm}
\captionof{figure}{Prompt used to summarize daily behavioral logs into intent-level memories.}
\vspace{0.5cm}
\label{fig:prompt_st}

\begin{tcolorbox}[width=\textwidth, colback=white, colframe=black, title=\texttt{\small{\# Persona Induction Prompt}}]
You are an analytical assistant that extracts “lifestyle-, personality-, and value-driven Personas” from user memories. \\
\small

    \textbf{Core Principles:}
    
    \hspace{0em} - A persona is not a simple topic or interest classification.

    \hspace{0em} - A persona must reveal the following elements together:
    
    \hspace{1em} 1) Repeated behavioral patterns (how the user searches for or consumes information)
    
    \hspace{1em} 2) Life context or rhythm (when, in what situations, and around what focus)
    
    \hspace{1em} 3) Personal tendencies or values (e.g., practical, emotional, experience-oriented)
    
    \hspace{0em} - Even if topics differ, group them into one persona if the exploration style and lifestyle pattern clearly overlap.
    
    \hspace{1em} - However, only group different topics together when the context clearly overlaps.
    
    \hspace{1em} - Avoid overly abstract or excessively broad personas.
    
    \hspace{0em} - A single log (memory\_id) may belong to multiple personas.
    
    \hspace{1em} - However, allow overlap only when the lifestyle context clearly intersects. \\
    
    \textbf{Input:}
    
    \hspace{0em} - A user’s search/recommendation service memories sorted in chronological order. \\
    
    \textbf{Task:}
    
    \hspace{0em} 1) Identify recurring patterns in how the user explores information and structures their daily life.
    
    \hspace{0em} 2) Based on these patterns, derive personas that reflect how this person spends their daily life and what they value.
    
    \hspace{1em} - There is no strict limit on the number of personas, but up to 10 is considered appropriate.
    
    \hspace{0em} 3) Each persona must include:
    
    \hspace{1em} - A concise noun phrase or descriptive phrase that conveys the lifestyle/tendency (persona)
    
    \hspace{1em} - 2–3 behavioral signatures that characterize the tendency (description)
    
    \hspace{0em} 4) Avoid exaggerated expressions.
    
    \hspace{0em} 5) Explicitly indicate which memory (memory\_id) support each persona. \\
    
    \hspace{0em} The output must strictly follow the JSON format below.
    
    \{
    
      \hspace{1em} "personas": [
      
        \hspace{2em} \{
        
          \hspace{3em} "persona\_id": "P1",
          
          \hspace{3em} "persona": "<One concise phrase describing the behavioral pattern.>",
          
          \hspace{3em} "description": ["<key behavior 1>",
            
            \hspace{9em} "<key decision criterion>",
            
            \hspace{9em} "<typical constraints>"],
          
          \hspace{3em} "evidence\_memory\_ids": [<memory\_id>, <memory\_id>, <memory\_id>, ...]
          
        \hspace{2em} \},
      
        \hspace{2em} ...
      
      \hspace{1em} ]
      
    \} \\
    
    \textbf{Memories:} \\
    \textcolor{gray}{\{\{ memories \}\}}
    \end{tcolorbox}
\vspace{-0.2cm}
\captionof{figure}{Prompt used for persona induction from intent-level memories.}
\vspace{0.5cm}
\label{fig:prompt_lt}

\begin{figure*}[h!]
    \small
    \begin{tcolorbox}[width=\textwidth, colback=white, colframe=black, title=\texttt{\# Judge Prompts: Alignment}]

    You are a strict evaluator assessing the alignment between a persona and its supporting memories. Below, one Persona description and multiple evidence memories are provided.
    \vspace{0.2cm}
    
    \textbf{Objective:}
    
    \hspace{1em} - Determine whether each evidence meaningfully aligns with the core characteristics of the Persona.
    
    \hspace{1em} - Count the number of evidence items that clearly align with the persona. 
    \vspace{0.1cm}

    \textbf{Evaluation Criteria:}

    \hspace{1em} - If an evidence item serves as direct support for the persona, count it as aligned.
    
    \hspace{1em} - If it is weakly related or reflects a different tendency, do not count it as aligned. 
    \vspace{0.1cm}
    
    \textbf{Score Calculation:}
    
    \hspace{1em} - score = (number of evidence items aligned with the persona) / (total number of evidence items)
    \vspace{0.2cm}
    
    Output must strictly follow the JSON format below:
    
    \{

      \hspace{1em} "score": <0.0--1.0>,
      
      \hspace{1em} "aligned\_evidence\_ids": [<int>, ...],
      
      \hspace{1em} "non\_aligned\_evidence\_ids": [<int>, ...]

    \}
    \vspace{0.2cm}
    
    \textbf{Input:} \\
    \textcolor{gray}{\{\{ $p_i$ \}\}}
    \end{tcolorbox}
\vspace{-0.2cm}
\caption{Prompt used for evaluating persona--memory alignment. The judge determines which evidence memories support the persona and computes an alignment score based on the proportion of aligned evidence.}
\vspace{0.5cm}
\label{fig:prompt_judge_alignment}
\end{figure*}

\begin{tcolorbox}[width=\textwidth, colback=white, colframe=black, title=\texttt{\small{\# Judge Prompts: Truthfulness}}]
    \small

    You are a highly strict evaluator of exaggeration or unsupported claims.
    \vspace{0.2cm}
    
    \textbf{Objective:}
    
    \hspace{1em} - For the single persona and for each description, determine whether they are supported by the provided evidence.
    
    \hspace{1em} - Any inference, speculation, or generalization not explicitly grounded in the evidence should be penalized. 
    \vspace{0.1cm}
    
    \textbf{Important Rules:}
    
    \hspace{1em} - Strong expressions such as “repeatedly,” or “always,” are considered exaggerated unless repeated patterns are 
    
    \hspace{1.5em} clearly confirmed in the evidence.
    
    \hspace{1em} - If there is little to no relevant Evidence, assign a low score (close to 0). 
    \vspace{0.1cm}
    
    \textbf{Scoring Guidelines (0.0–1.0):}
    
    \hspace{1em} - 1.0: Clearly and repeatedly supported by Evidence; no exaggeration.
    
    \hspace{1em} - 0.7: Mostly supported, but some wording is slightly strong (minor exaggeration).

    \hspace{1em} - 0.3: Partially supported. Some core claims lack evidence or are generalized.
    
    \hspace{1em} - 0.0: Little to no supporting evidence. Primarily speculative or exaggerated.
    \vspace{0.2cm}
    
    Output must strictly follow the JSON format below:
    
    \{

      \hspace{1em} "persona": \{
      
      \hspace{2em} "score": <0.0--1.0>, "overclaim\_phrases": ["<short phrase judged as exaggerated>", ...]
      
      \hspace{1em} \},
      
      \hspace{1em} "descriptions": [
      
      \hspace{2em} \{
        
          \hspace{3em} "id": 1,
          
          \hspace{3em} "score": <0.0--1.0>, "overclaim\_phrases": ["<short phrase judged as exaggerated>", ...]
          
      \hspace{2em} \},
        
      \hspace{2em} ...
        
    \hspace{1em} ]
    
    \}
    \vspace{0.2cm}
    
    \textbf{Input:} \\
    \textcolor{gray}{\{\{ $p_i$ \}\}}
    \end{tcolorbox}
\vspace{-0.2cm}
\captionof{figure}{Prompt used for evaluating persona truthfulness. The judge assesses whether persona descriptions are supported by the evidence and penalizes exaggerated or unsupported claims.}
\vspace{0.5cm}
\label{fig:prompt_judge_truth}

\begin{tcolorbox}[width=\textwidth, colback=white, colframe=black, title=\texttt{\small{\# PersonaX$_r$ Inference Prompt}}]
    \small
    \textbf{Task} \\
    We provide a user's personal profile in [User Profile], which includes the user's preferences, dislikes, and other relevant information. You need to play the role of the user. We also provide two candidate activities, A and B, with their descriptions in [Activity Description]. You need to choose between the two activity candidates based on your profile and the descriptions of the activities. Furthermore, you must articulate why you've chosen that particular activity while rejecting the other. \\

    \textbf{User Profile} \\
    \textcolor{gray}{\{\{ memory \}\}} \\

    \textbf{Activity Description} \\
    Activity A: \textcolor{gray}{\{\{ item\_a \}\}} \\
    Activity B: \textcolor{gray}{\{\{ item\_b \}\}} \\

    \textbf{Steps to Follow} \\
    1. Extract your preferences and dislikes from your profile. \\
    2. Evaluate the two candidates in light of your preferences and dislikes. Make your choice by considering the correlation between your preferences/dislikes and the features of the candidates. \\
    3. Explain why you made such choices, from the perspective of the relationship between your preferences/dislikes and the features of these candidate activities. \\

    \textbf{Important Notes} \\
    1. Your output should strictly be in the following format: \\
    \hspace{1em} Chosen Activity: \{Activity A or Activity B\} \\
    \hspace{1em} Explanation: \{Your detailed rationale ...\} \\
    2. When identifying user's likes and dislikes, do not fabricate them! If your [User Profile] doesn't specify any relevant preferences or dislikes, use common knowledge to inform your decision. \\
    3. You \textbf{must} choose one of these two candidates, and \textbf{cannot} choose both. \\
    4. Your explanation needs to be comprehensive and specific. \\
    5. Base your explanation on facts. \\
    6. Please ignore the effect of Activity position and length, they do not affect your decision.
\end{tcolorbox}
\vspace{-0.2cm}
\captionof{figure}{Inference prompt used in the $\text{PersonaX}_r$ (Reflection) pipeline. The model selects between a positive and negative activity pair based on its current persona profile.}
\vspace{0.5cm}
\label{fig:prompt_personax_a_infer}


\begin{tcolorbox}[breakable, width=\textwidth, colback=white, colframe=black, title=\texttt{\small{\# PersonaX$_r$ Reflection Prompt}}]
    \small
    \textbf{Background} \\
    We provide a user's personal profile in [User Profile], which includes the user's preferences, dislikes, and other relevant information. You need to play the role of the user. Recently, you considered choosing one more preferred activity from two candidates. The descriptions of these two candidates are provided in [Activity Description]. And your choice and explanation is in [Choice and Explanation], which reveals your previous judgment for these two candidates. \\

    \textbf{User Profile} \\
    \textcolor{gray}{\{\{ memory \}\}} \\

    \textbf{Activity Description} \\
    Activity A: \textcolor{gray}{\{\{ item\_a \}\}} \\
    Activity B: \textcolor{gray}{\{\{ item\_b \}\}} \\

    \textbf{Choice and Explanation} \\
    \textcolor{gray}{\{\{ response \}\}} \\

    \textbf{Task} \\
    However, the user in the real world actually prefers Activity B, and rejects Activity A that you initially chose. This indicates that you made an incorrect choice, the [Choice and Explanation] was mistaken. Therefore, you need to reflect and update [User Profile]. \\

    \textbf{Steps to Follow} \\
    1. Analyze the misconceptions in your previous [Choice and Explanation] about your preferences and dislikes, and correct these mistakes. \\
    2. Explore your new preferences based on Activity B you really enjoy, and determine your dislikes based on Activity A you truly don't enjoy. \\
    3. Summarize your past preferences and dislikes from your previous [User Profile]. Combine your newfound preferences and dislikes with your past ones. Filter and remove any conflicting or repetitive parts. \\
    4. Generate an updated profile as a JSON object with the format specified below. \\

    \textbf{Important Notes} \\
    - A persona is not a simple topic or interest classification. \\
    - A persona must reveal the following elements together: \\
    \hspace{1em} 1) Repeated behavioral patterns (how the user searches for and consumes information) \\
    \hspace{1em} 2) Life context or rhythm (when, in what situations, and around what focus) \\
    \hspace{1em} 3) Personal tendencies or values (e.g., practical, emotional, experience-oriented) \\
    - From the given logs, identify common exploration styles, lifestyle patterns, and emotional contexts to form a single persona. \\
    - Avoid overly abstract or excessively broad personas. Generate exactly one persona. \\
    - Only describe features of activities the user prefers or dislikes. Do not explain why the choice was wrong or why the profile was updated. \\

    The output must strictly follow the JSON format below: \\

    \{

      \hspace{1em} ``personas'': [

        \hspace{2em} \{

          \hspace{3em} ``persona\_id'': ``P1'',

          \hspace{3em} ``persona'': ``<One concise phrase describing the behavioral pattern.>'',

          \hspace{3em} ``description'': [``<key behavior 1>'',

            \hspace{9em} ``<key decision criterion>'',

            \hspace{9em} ``<typical constraints>'']

        \hspace{2em} \}

      \hspace{1em} ]

    \}
\end{tcolorbox}
\vspace{-0.2cm}
\captionof{figure}{Reflection prompt used in the $\text{PersonaX}_r$ (Reflection) pipeline. When the model makes an incorrect choice, this prompt instructs it to reflect on the mistake and update its persona profile.}
\label{fig:prompt_personax_a_reflect}

\end{document}